\documentclass[11pt]{article}
\PassOptionsToPackage{table}{xcolor}
\usepackage[final]{acl}
\usepackage{tabularx}
\usepackage{url}
\usepackage{booktabs}
\usepackage{graphicx}
\usepackage{booktabs}
\usepackage{multirow}
\usepackage{makecell}
\usepackage{xcolor}
\definecolor{deepPink}{RGB}{070,172,202}
\definecolor{lightPink}{RGB}{167,210,228}
\usepackage{arydshln}
\usepackage{amsmath,amssymb}
\DeclareMathSizes{11}{11}{8}{5}   
\usepackage[ruled,vlined,linesnumbered]{algorithm2e} 
\SetKwInput{Input}{Input}   
\SetKwInput{Output}{Output} 

\SetCommentSty{mycommentstyle} 
\usepackage{tcolorbox}
\usepackage{pifont}
\usepackage{kotex}
\tcbuselibrary{breakable,skins}
\tcbset{
  breakable,
  enhanced jigsaw,
  width=\columnwidth,
  left=2pt, right=2pt,
  fontupper=\small\raggedright,
}

\usepackage[T1]{fontenc}
\usepackage{lineno}
\definecolor{darkblue}{rgb}{0, 0, 0.5}
\usepackage{hyperref}
\newcommand\blfootnote[1]{\begingroup\renewcommand\thefootnote{}\footnote{#1}\addtocounter{footnote}{-1}\endgroup}
\hypersetup{colorlinks=true, citecolor=darkblue, linkcolor=darkblue, urlcolor=darkblue}
\usepackage{times}
\usepackage{latexsym}
\usepackage[T1]{fontenc}
\usepackage[utf8]{inputenc}
\usepackage{anyfontsize}
\usepackage{float}

\usepackage{inconsolata}

\usepackage{graphicx}

\usepackage{fancyhdr}
\newcommand{\aclfootershift}{-0.5pt}
\newcommand{\aclpagenumshift}{11.5pt}
\fancypagestyle{aclfirstpage}{%
  \fancyhf{}%
  \fancyfoot[C]{\raisebox{\aclfootershift}[0pt][0pt]{\parbox{\textwidth}{\centering
    {\normalsize\thepage\par}\vspace{3pt}
    {\footnotesize\setlength{\baselineskip}{11pt}%
     \textit{Proceedings of the 2026 Conference on Empirical Methods in Natural Language Processing}, pages 11523--11554\\
     October 24-29, 2026 \textcopyright 2026 Association for Computational Linguistics\par}}}}%
}
\fancypagestyle{aclotherpages}{%
  \fancyhf{}%
  \fancyfoot[C]{\raisebox{\aclpagenumshift}[0pt][0pt]{\normalsize\thepage}}%
}

\title{Correct Prediction, Wrong Steps? Consensus Reasoning Knowledge Graph for Robust Chain-of-Thought Synthesis}

\author{
  Zipeng Ling$^{1,2,3}$\thanks{Equal Contribution},
  Shuliang Liu$^{1}$\footnotemark[1],
  Seonil Son$^{4}$,
  Shenghong Fu$^{5}$,
  \\
  \textbf{
  Yuehao Tang$^{1}$,
  Yao Wan$^{6}$,
  Xuming Hu$^{1}$}\thanks{Corresponding Author} \\
  $^{1}$The Hong Kong University of Science and Technology (Guangzhou)\\
  $^{2}$University of Alberta
  $^{3}$Alberta Machine Intelligence Institute $^{4}$RLWRLD\\ $^{5}$The Hong Kong Polytechnic University\\
  $^{6}$Huazhong University of Science and Technology\\
  \texttt{\{zpling0816, xuminghu97\}@gmail.com}
}

\begin{document}
\setcounter{page}{11523}
\pagestyle{aclotherpages}
\maketitle
\thispagestyle{aclfirstpage}

\begin{abstract}
{
Large language models (LLMs) have become increasingly used for various tasks, often coupled with Chain-of-Thought (CoT) prompting to boost accuracy. Recent work has shown that high label-prediction accuracy does not guarantee correct intermediate reasoning, and the causes of \textit{reasoning flaws} vary from sample to sample, yet existing remedies either focus on a single domain or assume that one flaw type applies uniformly across samples. A simple mitigation method is to provide the model with the correct answer, but we show that this yields no consistent improvement in reasoning quality. This indicates that the problem cannot be fixed by LLMs' awareness of answers, and must instead be addressed through the \textit{structure} of reasoning. Motivated by this, we propose \textsc{CRAFT} (Consensus Reasoning knowledge graph Aggregation for Flaw-aware Trace synthesis), which aggregates the consensus components shared across multiple candidate reasoning traces to synthesize improved ones. \textsc{CRAFT} consistently improves label-prediction accuracy on both logical and mathematical reasoning benchmarks, outperforming most baselines, while its post-processed traces achieve higher quality under fine-grained benchmark evaluation.
}
\end{abstract}

\section{Introduction}
\label{Introduction}

\blfootnote{Code and datasets are available at the \href{https://github.com/ZpLing/CRAFT}{GitHub repository}.}LLMs show impressive reasoning ability in logical~\citep{zhou2025dissectinglogicalreasoningllms} and mathematical domains~\citep{shao2024deepseekmathpushinglimitsmathematical}, increasingly trained via Reinforcement Learning that distinguishes good traces from bad~\citep{lightman2023letsverifystepstep}.
These capabilities have been adopted across high-stakes domains like financial analysis~\citep{zhao2024financemathknowledgeintensivemathreasoning}, medical treatment~\citep{savage2023diagnosticreasoningpromptsreveal}, and legal consultation~\citep{yu2022legalpromptingteachinglanguage}.
In such applications, what matters is not only the final answer an LLM outputs, i.e., its label-prediction accuracy, but also the quality of its reasoning trace, i.e., whether the intermediate steps leading to that answer are themselves logical and sound. Conventionally, the latter is taken for granted: a correct answer is read as evidence of correct reasoning.

\begin{figure}[!t]
  \centering

  \includegraphics[
    width=\columnwidth,
      keepaspectratio,
      trim=0 00pt 0 0,
      clip
  ]{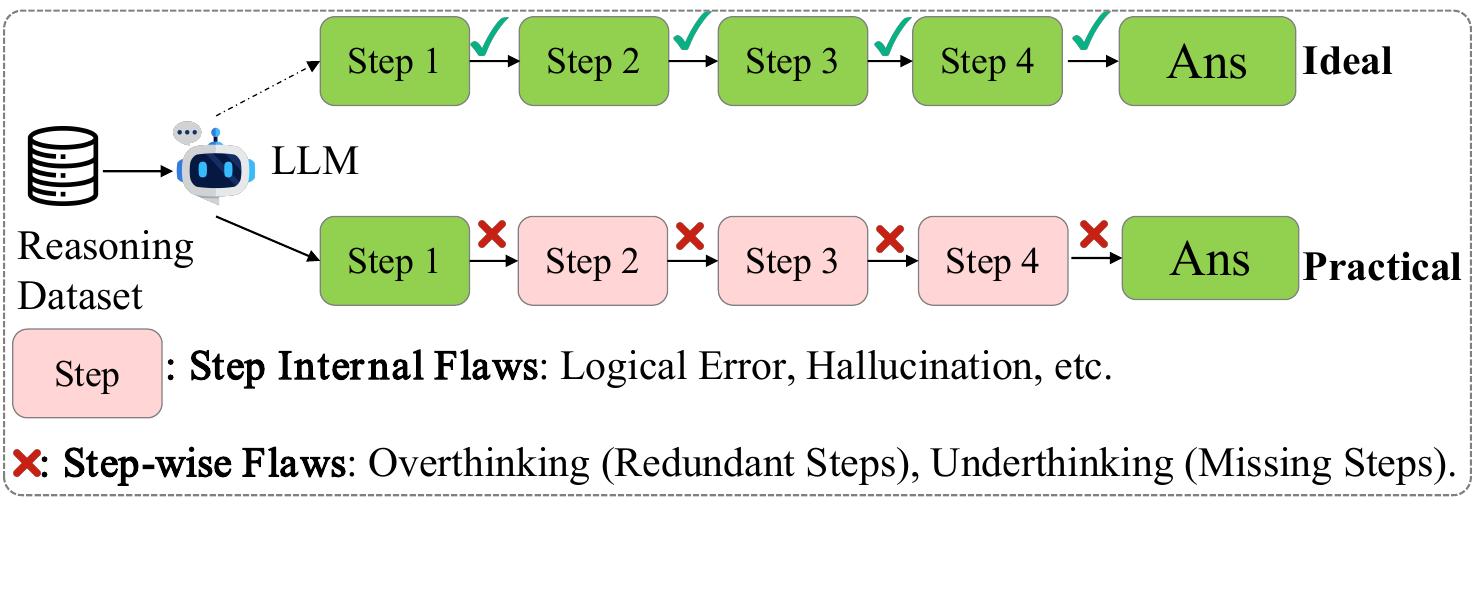}
  \vspace{-30pt}
  \caption{Problem background. LLMs' correct label-prediction is related to, but not fully determined by, whether reasoning steps are correct. Case studies of logical and mathematical reasoning are in Appendix~\ref{app:case_study}.
  }
  \label{fig:problem background}
\end{figure}

However, this assumption does not always hold: correct label-prediction is necessary but not sufficient evidence of correct reasoning. For instance, self-reflection based on previously generated traces can result in accuracy drops~\citep{ling2025instructionboundaryquantifyingbiases,ling2025wakenllmevaluatingreasoningpotential}, and by optimizing only for the final answer, RL can favor reasoning steps that reach the correct prediction through invalid intermediate logic~\citep{yue2025doesreinforcementlearningreally}.
Figure~\ref{fig:problem background} visualizes this: whereas an ideal trace has every step correct, in practice a trace can reach the correct answer despite flawed intermediate steps. Such flaws fall into two types: \emph{Step Internal Flaws}, where an individual step is itself wrong (e.g., a logical error or hallucination), and \emph{Step-wise Flaws}, where steps may each look valid but are mis-organized: Omitting necessary steps (Underthinking) or piling on redundant ones (Overthinking).
This poses risks for (1) {LLM Distillation}: models may be fine-tuned on traces generated by other LLMs; 
and (2) {Dataset Annotation}: such errors may propagate into benchmarks that rely on LLM-generated traces to annotate problem-solving methods.

Existing mitigation efforts fall short on two aspects.
\textbf{(i)~Reasoning Domain coverage:} ~\citet{xu2026logicrewardincentivizingllmreasoning} only targets the problem in Logical Reasoning domain. As their method is tailored to the domain specifically, it does not transfer to other domains such as mathematical reasoning, even though the same reasoning flaws arise there. \textbf{(ii)~Flaw-type coverage:} each method assumes a single flaw type holds for all traces: some attribute flaws to redundant steps (Overthinking)~\citep{yu2025causalsufficiencynecessityimproves}, others to missing steps (Underthinking)~\citep{xu2025mindgapbridgingthought}, yet in practice different samples trigger different flaws, yielding a complex mixture that no single-type assumption can fully capture.

Before proposing a problem-mitigation method, we first ask why these reasoning flaws arise. A natural hypothesis is that they arise from the model's uncertainty about the correct answer; if so, the simplest fix would be to provide the answer and ask the model merely to justify it. We test this hypothesis on two benchmarks, (1) PRMBench~\cite{song2025prmbenchfinegrainedchallengingbenchmark}: evaluating an LLM as a step-by-step verifier for detecting error steps, and (2) ROSCOE~\citep{golovneva2023roscoesuitemetricsscoring}: evaluating reasoning-trace quality.
Across four LLMs, providing the correct answer does not improve reasoning in most settings (Section~\ref{sec:label_guidance_results}), indicating that the flaws are structural rather than a matter of answer-awareness.

Since LLMs' awareness of answers does not help, the flaws lie in the reasoning trace itself, not the final answer. Our key observation is that flawed steps tend to be sample-specific, whereas correct ones recur across independently sampled traces~\cite{wang2023selfconsistencyimproveschainthought}. This motivates mitigating both flaw types via \emph{cross-trace consensus}: aggregating correct reasoning steps of multiple candidate traces, while flawed ones are dropped out. To apply consensus to step content (catching \emph{Step Internal Flaws}) and step structure (catching \emph{Step-wise Flaws}), we propose \textbf{CRAFT} (Consensus Reasoning knowledge graph Aggregation for Flaw-aware Trace synthesis), a unified framework built on a \emph{Reasoning Knowledge Graph} (RKG) that operates in three modules:
\textbf{(I)~Multi-Trace Generation \& Steps Filtering}: roll out $K$ candidate traces per sample, identify consensus terms via TF-IRF weighting, and remove structurally different steps through z-score filtering;
\textbf{(II)~Consensus RKG Construction}: convert traces into per-trace RKGs (steps as nodes, logical relations as edges), aggregate them into a consensus RKG $G^{*}$, and filter minor edges and nodes;
\textbf{(III)~Topology-guided Trace Synthesis}: generate one high-quality trace by traversing $G^{*}$ in topological order.

To summarize, our contributions are three-fold:

\noindent$\triangleright$ A benchmark study showing that providing correct answers yields no consistent improvement in reasoning, which indicates that the flaws are structural rather than a matter of answer-awareness.

\noindent$\triangleright$ We propose CRAFT, a unified framework that jointly mitigates \emph{Step Internal Flaws} and \emph{Step-wise Flaws}, yielding consistent label-prediction accuracy gains across both logical and mathematical reasoning datasets.

\noindent$\triangleright$ We further show that CRAFT's post-processed traces achieve higher quality under fine-grained benchmark evaluation.



\section{Related Work}

\paragraph{LLM Reasoning Flaws.}
Chain-of-Thought prompting~\cite{wei2023chainofthoughtpromptingelicitsreasoning} makes intermediate steps explicit, but generated traces suffer from two flaw categories: \emph{Step Internal Flaws} (logical errors~\cite{xu2024largelanguagemodelsreally}, hallucinations~\cite{xia2025evaluatingmathematicalreasoningaccuracy}, etc.) and \emph{Step-wise Flaws} (underthinking~\cite{xu2025mindgapbridgingthought} and overthinking~\cite{yu2025causalsufficiencynecessityimproves}). The single-flaw assumption cannot describe practical scenarios, where different samples exhibit different flaw types.

\paragraph{Reasoning Traces Improvement.}
Recursive Self-Aggregation~\cite{venkatraman2026recursiveselfaggregationunlocksdeep} iteratively combine subsets of intermediate steps across candidate chains, exploiting intra-chain structure, but without an explicit anomaly filtering. Multi-chain graph methods such as MGRS~\cite{yang2025mgrs} construct a graph over multiple chains, but ultimately select the highest-scoring trace rather than making use of others. CRAFT applies cross-trace consensus and anomaly detection to mitigate different flaw types. Detailed related work is available in Appendix~\ref{Detailed Related Work}.

\begin{figure*}[t]
  \centering
  \includegraphics[
      width=\textwidth,
      keepaspectratio,
      trim=0 0 0 0,
      clip
  ]{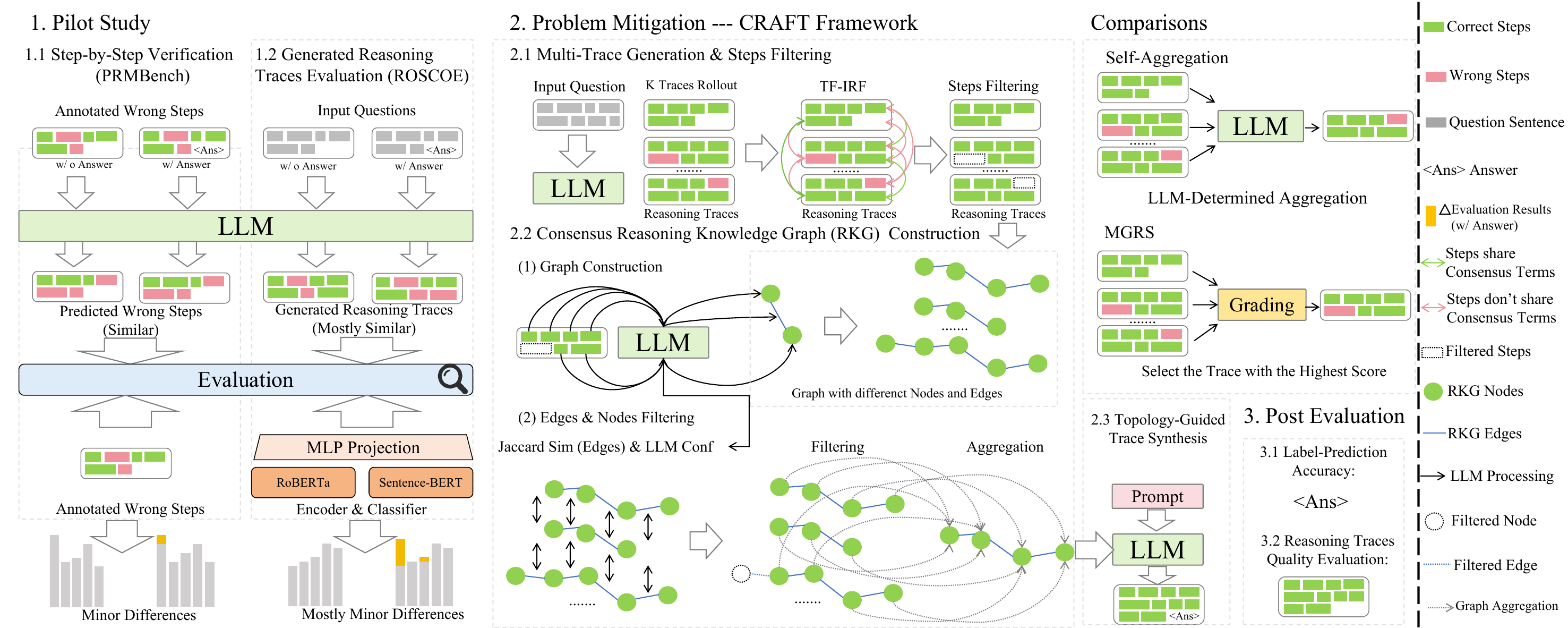}
  \caption{Overview of our work. \textbf{First}, we show that providing the correct answer does not consistently improve LLM reasoning ability, which motivates CRAFT, a structural approach for problem mitigation. \textbf{Second}, we show how CRAFT operates within the framework. \textbf{Third}, we test label-prediction accuracy on both logical and mathematical reasoning datasets, and post-processed reasoning traces on fine-grained benchmark evaluation.}
  \label{fig:Main Experiments}
  \vspace{-10pt}
\end{figure*}

\section{Methodology}
\label{sec:method}

\subsection{Pilot Study}
\label{sec:benchmark_eval}

A natural first question is whether flaws in LLM reasoning arise from uncertainty about the correct answer. If so, the simplest fix would be to provide the answer directly, and merely require the model to justify the given answer.
We test this hypothesis on two complementary benchmarks: (1) PRMBench~\cite{song2025prmbenchfinegrainedchallengingbenchmark}: benchmarking an LLM as a step-by-step verifier for identifying erroneous steps, and (2) ROSCOE~\citep{golovneva2023roscoesuitemetricsscoring}: benchmarking reasoning trace quality generated by an LLM. Experiments are carried out across two settings: (1) \textbf{w/ Answer}: the correct answer is provided in the prompt, and (2) \textbf{w/o Answer}: no correct answer is provided.

The results (Section~\ref{sec:label_guidance_results}) are counterintuitive: In general, providing the correct answer yields no consistent improvement in reasoning. This rules out answer uncertainty as the cause of reasoning flaws, pointing to the issue of reasoning traces structure. Also, when LLMs reason over a dataset, the result can be a mixture of flaw types: some reasoning suffer from overthinking, others from underthinking, and still others from \emph{Step Internal Flaws}, all generated within the same dataset. A framework must mitigate both \emph{Step Internal Flaws} and \emph{Step-wise Flaws}.


\subsection{Proposed Framework: CRAFT}

To mitigate this, we propose \textbf{CRAFT} (Consensus Reasoning knowledge graph Aggregation for Flaw-aware Trace synthesis), a unified framework that mitigates both \emph{Step Internal Flaws} (logical errors, hallucinations, etc.) and \emph{Step-wise Flaws} (overthinking, underthinking). CRAFT rolls out $K$ diverse candidate traces per sample and leverages cross-trace consensus to detect both flaw types: steps with erroneous content fail to appear across other traces (Step Internal Flaws), while missing or redundant steps are revealed by comparing each trace's structure against others (Step-wise Flaws). Inspired by GRPO's group relative comparison~\cite{Guo_2025}, we identify flawed steps as deviations from intra-group consensus.

Inspired by Knowledge Graphs~\cite{Besta_2024}, we construct a \emph{Reasoning Knowledge Graph} (RKG) in which nodes are reasoning steps and edges are logical relations, and RKGs $G_S$ are aggregated into a consensus RKG $G^{*}$ whose topology guides trace synthesis. CRAFT proceeds in three modules (Algorithm~\ref{Algorithms}):
\textbf{Module~I} (Multi-Trace Generation \& Steps Filtering) rolls out $K$ candidate traces and extracts consensus terms $T_{\text{Con}}$ via TF-IRF, and removes flawed steps via z-score filtering.
\textbf{Module~II} (Consensus RKG Construction) constructs a reasoning knowledge graph (RKG) for each reasoning trace, aggregates them into a consensus RKG $G^{*}$, and applies edge \& node filtering;
\textbf{Module~III} (Topology-guided Trace Synthesis) generates one high-quality trace by traversing $G^{*}$ in topological order.

\begin{algorithm*}[!t]
\caption{CRAFT for Reasoning Trace Improvement (One Sample)}
\label{Algorithms}
\fontsize{11pt}{12pt}\selectfont
\tcp{\textbf{Module~I: Multi-Trace Generation \& Steps Filtering}}
$\mathcal{C} \leftarrow \emptyset$\;
\For{$i = 1$ \textbf{to} $K$}{$\mathcal{C} \leftarrow \mathcal{C} \cup \{\textsc{LLM}(Sample)\}$\;}
$T_{\text{Con}} \leftarrow \left\{ w \mid \mathrm{TF\text{-}IRF}(w) > \alpha \;\wedge\; \overline{\mathrm{RF}}(w) \ge \beta \right\}$\tcp*{step-level consensus terms}

\ForEach{step $s$ of trace $S \in \mathcal{C}$}{
  $Z(s) \leftarrow (\mathrm{Jaccard}(s,T_{\text{Con}}) - \mu)/\sigma$\tcp*{Jaccard similarity, z-score steps filtering}
  \lIf{$Z(s) < \gamma$}{remove $s$ from $S$}
}
\BlankLine
\tcp{\textbf{Module~II: Consensus RKG Construction}}
\tcp{(1) Graph Construction}
\ForEach{trace $S \in \mathcal{C}$}{$G_S=(V_S,E_S) \leftarrow \textsc{LLM}(S)$\tcp*{Build RKG, steps$\to$nodes, logical relations$\to$edges}}
\tcp{(2) Edges \& Nodes Filtering }
\ForEach{trace $S \in \mathcal{C}$ \textbf{and} $e{=}(u,v) \in E_S$}{
  $W_S(e) \leftarrow (1{-}\lambda) \cdot \mathrm{conf}(e) + \lambda \cdot \mathrm{Jaccard}(u,v)$\tcp*{edge weight: LLM conf + node Jaccard}
}
$G^*=(V^*,E^*) \leftarrow \bigcup_{S \in \mathcal{C}} G_S$\tcp*{aggregation into one graph}
\ForEach{$e \in E^*$}{$W(e) \leftarrow \tfrac{1}{K}\sum_{S \in \mathcal{C}:\, e \in E_S} W_S(e)$\tcp*{average edges weight}}
$G^* \leftarrow G^* \setminus \{e : W(e){<}\theta\} \setminus \{v:d^{\!+}(v){=}0\wedge d^{\!-}(v){=}0\}$\tcp*{edges \& nodes filtering}
\BlankLine
\tcp{\textbf{Module~III: Topology-guided Trace Synthesis}}
\While{$\exists\, c=\{(v_1,v_2),\ldots,(v_{m-1},v_m),(v_m,v_1)\}\subseteq E^{*}$}{
  $E^{*} \leftarrow E^{*} \setminus \{\arg\min_{e \in c} W(e)\}$\tcp*{break graph cycles}
}
$\pi \leftarrow (v_1, \ldots, v_{|V^{*}|})\ \text{s.t.}\ (v_i, v_j) \in E^{*} \Rightarrow i < j$\tcp*{topological sort over $G^{*}$}
$\mathcal{T}_{\text{Final}} \leftarrow [\,]$ \;
\ForEach{node $v$ in $\pi$}{
  $s(v) \leftarrow \textsc{LLM}\!\left(Sample,\; \{s(u) : (u,v) \in E^{*}\},\; v\right)$\tcp*{step-by-step generation}
  Append $s(v)$ to $\mathcal{T}_{\text{Final}}$\;
}
\Return{$\mathcal{T}_{\text{Final}}$}\;
\end{algorithm*}

\paragraph{Module~I: Multi-Trace Generation \& Steps Filtering}
For each sample, we roll out $K$ candidate traces at temperature $T$.
We score each term by \emph{TF-IRF} (Term Frequency--Inverse Reasoning Frequency), a variant of TF-IDF~\cite{das2023comparativestudytfidffeature} adapted to the multi-trace setting: within one sample's $K$ traces, terms shared by many steps form the consensus, while a term that appears in only a few steps marks those steps as potential outliers.
Both TF and IRF are computed within the \emph{K} traces of one sample, the formula is available at Appendix~\ref{app:tfirf}, Eqs.~\ref{eq:tf}--\ref{eq:tfirf}). After filtering common logical connectives via a \textsc{CommonLogicalWords} blocklist~\citep{bird-etal-2009-nltk}, the Consensus Terms Set $T_{\text{Con}}$ is defined directly by combining a per-term importance threshold and a cross-trace frequency threshold:
$T_{\text{Con}}=\{w\mid\mathrm{TF\text{-}IRF}(w)>\alpha \;\wedge\; \overline{\mathrm{RF}}(w)\ge\beta\}$, where $\alpha$ is the TF-IRF threshold and $\beta$ is the step frequency threshold (fraction of the $K$ traces' steps in which a term must appear, Eq.~\ref{eq:rf}), and $T_{\text{Con}}$ guides steps filtering.

For Steps Filtering, following GRPO's group-relative comparison~\cite{Guo_2025}, each step $s$ is scored by its terms' Jaccard similarity with $T_{\text{Con}}$, z-normalized over all steps of the $K$ traces. Steps whose z-score falls below a cutoff $\gamma$ (default $-1.0$, i.e., more than one standard deviation below the group mean) are removed as outliers, catching \emph{Step Internal Flaws} (erroneous steps introduce terms absent from other $K-1$ traces) and \emph{Step-wise Flaws} (redundant overthinking steps carry unusual terms).


\paragraph{Module~II: Consensus RKG Construction}
This module constructs the consensus Reasoning Knowledge Graph (RKG) based on two parts: 

(1) Graph Construction. After steps filtering, each trace is converted into an RKG $G_S$, via an LLM prompt that converts steps as nodes and logical relationships as edges.

(2) Edges \& Nodes Filtering. In each $G_S$, every edge gets a weight $W_S(e)$ that combines the LLM-reported confidence score with the Jaccard similarity between the two consecutive steps. Per-trace graphs $G_S$ are aggregated into one consensus RKG $G^{*}$ by taking the set union. During Graph Construction in (1), nodes and edges produced by LLMs are identical when they correspond to identical steps and logical relationships; therefore, set aggregation of graphs introduces no redundant or repetitive elements. Each edge of $G^{*}$ then receives the consensus weight $W(e)$, the sum of its $W_S(e)$ over the traces that contain it divided by $K$, so a trace without this edge contributes zero. Edges whose $W(e)$ is below $\theta$ (the edge filtering threshold) are filtered, ensuring the graph retains relations supported by multiple candidate traces. We then apply node filtering to $G^{*}$, removing isolated nodes whose in-degree and out-degree both equal zero, i.e., $d^{\!+}(v){=}0 \wedge d^{\!-}(v){=}0$. Detailed filtering statistics are in Tables~\ref{tab:zscore_stats} and~\ref{tab:rkg_stats}.

\noindent\textbf{Domain-specific Optimization.} For logical reasoning, steps expressing the same inference with different words receive high LLM confidence, down-weighting Jaccard influence. For mathematical reasoning, the LLM-confidence component of $W(e)$ further handles cases where equivalent expressions appear in different forms, and provably incorrect calculations are removed.

\paragraph{Module~III: Topology-guided Trace Synthesis} $G^{*}$ merges the RKGs of $K$ traces, and we perform a topological sort of $G^{*}$, first removing the lowest-$W(e)$ edge of any cycle left in the graph, and generate the final reasoning trace step by step. Each step's prompt includes the corresponding node from the consensus RKG. Prompt templates are available in Appendix~\ref{Prompt Templates}.

\paragraph{Hyperparameter Settings.}
Hyperparameters of Algorithm~\ref{Algorithms} are shown in Table~\ref{tab:hyperparams}, they were set once and not changed, reducing overfitting risks.

\begin{table}[H]
  \centering
  \renewcommand{\arraystretch}{1}
  {\small
  \resizebox{1.0\linewidth}{!}{%
  \begin{tabular}{llcl}
  \toprule
  \textbf{Symbol} & \textbf{Parameter} & \textbf{Value} &
  \textbf{Module} \\
  \midrule
  $K$       & Number of rollout traces        & 5      & I  \\
  $\alpha$  & TF-IRF threshold & 0.01   & I \\
  $\beta$   & Step frequency threshold & 0.3 & I  \\
  $\gamma$  & Z-score step cutoff     & $-1.0$ & I  \\
  $\lambda$ & Weight of Jaccard in $W_S(e)$ & 0.3 & II \\
  $\theta$  & Edge filtering threshold   & 0.3    & II \\
  \bottomrule
  \end{tabular}%
  }}
  \vspace{-6pt}
  \caption{Hyperparameters of Algorithm~\ref{Algorithms}.}
  \vspace{-6pt}
  \label{tab:hyperparams}
\end{table}

For models whose sampling temperature can be set, the $K{=}5$ rollout traces are sampled at temperatures 0.4, 0.55, 0.7, 0.85 and 1.0; Models do not expose a temperature are sampled at their default.

\subsection{Mitigation of Different Flaw Types}
\label{sec:flaws_paradigm}
The principle of flaw mitigation is a simple mechanism: steps shared by many candidate traces are preserved, while those shared by fewer are filtered.

\noindent\textbf{Step Internal Flaws} (logical errors, hallucinations, etc.). Error terms are absent from most candidate traces, so their steps score below $\gamma$ under Module~I's z-score filtering and are filtered.

\noindent\textbf{Step-wise Flaws.} \emph{Underthinking}: steps missing from one trace are preserved by others in $G^{*}$. \emph{Overthinking}: redundant steps carry unusual terms (low $Z$) and participate in low-weight edges (low $W(e)$), so they can be removed by both z-score filtering in Module~I and edge filtering in Module~II.


\begin{table*}[!t]
\centering
\resizebox{\textwidth}{!}{%
\renewcommand{\arraystretch}{1.05}
{\fontsize{7}{9}\selectfont
\begin{tabular}{ll l ccc @{\hspace{3pt}\vrule width 0.4pt\hspace{3pt}} l cccc}
\toprule
& & \multicolumn{4}{c}{\textbf{PRMBench}} & \multicolumn{5}{c}{\textbf{ROSCOE}} \\
\cmidrule(lr){3-6}\cmidrule(lr){7-11}
\textbf{Model} & \textbf{Setting} & \textbf{Dimension} & \textbf{StepAcc} & \textbf{1stErr} & \textbf{F1} & \textbf{Dataset} & \textbf{Faithfulness} & \textbf{Info (Step)} & \textbf{Info (Chain)} & \textbf{Grammar} \\
\midrule
\multirow{8}{*}{\shortstack{o4-mini}}
  & \multirow{4}{*}{w/ Answer}
    & Simplicity & 0.72 & 0.65 & 0.82 & CosmosQA & 0.82 & 0.81 & 0.92 & 0.95 \\
  & & Soundness & 0.85 & 0.74 & 0.91 & DROP & 0.83 & 0.81 & 0.92 & 0.96 \\
  & & Sensitivity & 0.85 & 0.61 & 0.91 & eSNLI & 0.71 & 0.78 & 0.87 & 0.83 \\
  & & Total & 0.81 & 0.67 & 0.88 & GSM8K & 0.82 & 0.84 & 0.96 & 0.94 \\
\cmidrule(lr){2-11}
  & \multirow{4}{*}{w/o Answer}
    & Simplicity & 0.74 & 0.64 & 0.83 & CosmosQA & 0.80 & 0.79 & 0.92 & 0.96 \\
  & & Soundness & 0.84 & 0.77 & 0.90 & DROP & 0.83 & 0.81 & 0.93 & 0.96 \\
  & & Sensitivity & 0.86 & 0.59 & 0.92 & eSNLI & 0.71 & 0.78 & 0.87 & 0.84 \\
  & & Total & 0.81 & 0.66 & 0.88 & GSM8K & 0.83 & 0.85 & 0.96 & 0.95 \\
\midrule
\multirow{8}{*}{\shortstack{GPT-5.4-nano}}
  & \multirow{4}{*}{w/ Answer}
    & Simplicity & 0.62 & 0.64 & 0.72 & CosmosQA & 0.83 & 0.81 & 0.91 & 0.91 \\
  & & Soundness & 0.69 & 0.76 & 0.79 & DROP & 0.84 & 0.81 & 0.92 & 0.94 \\
  & & Sensitivity & 0.75 & 0.47 & 0.85 & eSNLI & 0.69 & 0.78 & 0.86 & 0.84 \\
  & & Total & 0.68 & 0.62 & 0.78 & GSM8K & 0.81 & 0.82 & 0.96 & 0.94 \\
\cmidrule(lr){2-11}
  & \multirow{4}{*}{w/o Answer}
    & Simplicity & 0.67 & 0.57 & 0.77 & CosmosQA & 0.82 & 0.79 & 0.92 & 0.91 \\
  & & Soundness & 0.70 & 0.75 & 0.80 & DROP & 0.84 & 0.80 & 0.92 & 0.95 \\
  & & Sensitivity & 0.75 & 0.50 & 0.85 & eSNLI & 0.69 & 0.78 & 0.86 & 0.84 \\
  & & Total & 0.70 & 0.61 & 0.81 & GSM8K & 0.81 & 0.83 & 0.96 & 0.94 \\
\midrule
\multirow{8}{*}{\shortstack{DeepSeek-V4-Flash}}
  & \multirow{4}{*}{w/ Answer}
    & Simplicity & 0.82 & 0.47 & 0.89 & CosmosQA & 0.84 & 0.81 & 0.91 & 0.92 \\
  & & Soundness & 0.84 & 0.73 & 0.90 & DROP & 0.83 & 0.80 & 0.92 & 0.96 \\
  & & Sensitivity & 0.85 & 0.58 & 0.91 & eSNLI & 0.69 & 0.77 & 0.86 & 0.85 \\
  & & Total & 0.83 & 0.59 & 0.90 & GSM8K & 0.82 & 0.84 & 0.96 & 0.95 \\
\cmidrule(lr){2-11}
  & \multirow{4}{*}{w/o Answer}
    & Simplicity & 0.77 & 0.50 & 0.86 & CosmosQA & 0.82 & 0.79 & 0.92 & 0.94 \\
  & & Soundness & 0.84 & 0.71 & 0.90 & DROP & 0.83 & 0.80 & 0.92 & 0.96 \\
  & & Sensitivity & 0.84 & 0.54 & 0.91 & eSNLI & 0.68 & 0.77 & 0.86 & 0.85 \\
  & & Total & 0.82 & 0.58 & 0.89 & GSM8K & 0.82 & 0.84 & 0.96 & 0.95 \\
\midrule
\multirow{8}{*}{\shortstack{Gemini-3.1-Flash-Lite}}
  & \multirow{4}{*}{w/ Answer}
    & Simplicity & 0.73 & 0.69 & 0.82 & CosmosQA & 0.82 & 0.79 & 0.91 & 0.96 \\
  & & Soundness & 0.84 & 0.72 & 0.91 & DROP & 0.83 & 0.80 & 0.92 & 0.95 \\
  & & Sensitivity & 0.84 & 0.47 & 0.91 & eSNLI & 0.68 & 0.76 & 0.86 & 0.87 \\
  & & Total & 0.80 & 0.63 & 0.88 & GSM8K & 0.82 & 0.83 & 0.95 & 0.94 \\
\cmidrule(lr){2-11}
  & \multirow{4}{*}{w/o Answer}
    & Simplicity & 0.71 & 0.72 & 0.80 & CosmosQA & 0.80 & 0.78 & 0.91 & 0.96 \\
  & & Soundness & 0.84 & 0.71 & 0.90 & DROP & 0.83 & 0.80 & 0.92 & 0.95 \\
  & & Sensitivity & 0.83 & 0.48 & 0.90 & eSNLI & 0.69 & 0.76 & 0.87 & 0.87 \\
  & & Total & 0.79 & 0.64 & 0.87 & GSM8K & 0.83 & 0.84 & 0.95 & 0.96 \\
\bottomrule
\end{tabular}}}
\vspace{-6pt}
\caption{Pilot Study: \textbf{PRMBench} step-level verifier (\emph{Dimension} = PRMBench error dimension; StepAcc, 1stErr, F1) and \textbf{ROSCOE} trace quality (four metrics applicable to all datasets).}
\vspace{-10pt}
\label{tab:combined_eval}
\end{table*}

\begin{figure*}[t]
  \centering
  \includegraphics[width=\textwidth,keepaspectratio]{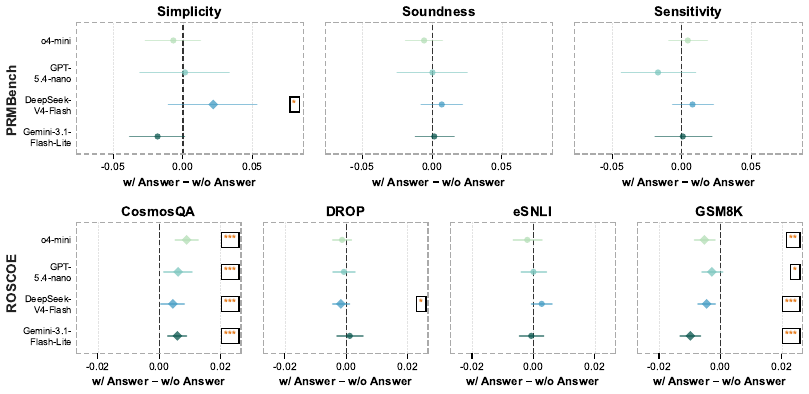}
  \vspace{-25pt}
  \caption{Significance testing (paired Wilcoxon, \textbf{w/ Answer} minus \textbf{w/o Answer}, 95\% bootstrap CI) on PRMBench (top) and ROSCOE (bottom). {\color[HTML]{E67E22}$*$}: $p<0.05$, {\color[HTML]{E67E22}$**$}: $p<0.01$, {\color[HTML]{E67E22}$***$}: $p<0.001$.}
  \label{fig:significance_forest}
\vspace{-10pt}
\end{figure*}

\section{Experiments}
\label{sec:experiments}

\subsection{Experimental Setup of Pilot Study}
\label{sec:label_guidance_results}

We first validate the hypothesis: does providing correct answers to questions improve LLMs' reasoning ability? We evaluate this using two benchmarks:
(1) \textbf{PRMBench}~\cite{song2025prmbenchfinegrainedchallengingbenchmark} benchmarks an LLM as a step-by-step verifier to detect erroneous steps across three error dimensions: \emph{Simplicity} (redundant steps), \emph{Soundness} (logically invalid steps), and \emph{Sensitivity} (missing conditions, deception and multiple solutions), with 200 items each. Metrics include step accuracy (StepAcc), F1, and first-error identification accuracy (1stErr).
(2) \textbf{ROSCOE}~\citep{golovneva2023roscoesuitemetricsscoring} evaluates the quality of reasoning traces generated by the tested LLM. We test this on 125 problems from each of four datasets: CosmosQA, DROP, eSNLI, GSM8K, measuring \emph{Faithfulness}, step-level and chain-level \emph{Informativeness}, and \emph{Grammar}.

We evaluate four backbone LLMs: o4-mini, GPT-5.4-nano, DeepSeek-V4-Flash, and Gemini-3.1-Flash-Lite. All experiments are conducted with the same prompt templates (Appendix~\ref{app:prmbench_prompts}), under two experimental settings: (1) \textbf{w/ Answer}: the correct answer is provided in the prompt, and (2) \textbf{w/o Answer}: no correct answer is provided.

\begin{table*}[!t]
  \centering
  \resizebox{\textwidth}{!}{%
  \renewcommand{\arraystretch}{1}
  {\fontsize{7}{9}\selectfont
  \begin{tabular}{l ccc!{\vrule width 0.4pt height 1.9ex depth 0.6ex} ccc!{\vrule width 0.4pt height 1.9ex depth 0.6ex} cc!{\vrule width 0.4pt height 1.9ex depth 0.6ex} cc}
  \toprule
  & \multicolumn{3}{c}{\textbf{FLD}}
  & \multicolumn{3}{c}{\textbf{ProofWriter}}
  & \multicolumn{2}{c}{\textbf{Omni-MATH}}
  & \multicolumn{2}{c}{\textbf{OlympiadBench}} \\
  \cmidrule(lr){2-4}\cmidrule(lr){5-7}\cmidrule(lr){8-9}\cmidrule(lr){10-11}
  \textbf{Setting}
  & Acc(\%)$\uparrow$ & F1$\uparrow$ & Steps$\downarrow$
  & Acc(\%)$\uparrow$ & F1$\uparrow$ & Steps$\downarrow$
  & Acc(\%)$\uparrow$ & Steps$\downarrow$
  & Acc(\%)$\uparrow$ & Steps$\downarrow$ \\
  \midrule
  \rowcolor{gray!20}\multicolumn{11}{c}{\textbf{GPT-5.4-nano}} \\
  \midrule
      Direct
      & 66.1 & 0.66 & -- & 72.0 & 0.71 & -- & 6.9 & -- & 4.0 & -- \\
      Raw CoT~\cite{wei2023chainofthoughtpromptingelicitsreasoning}
      & 77.0 & 0.77 & 15.7 & 84.6 & 0.84 & 22.9 & 33.7 & 40.4 & 44.1 & 40.2 \\
      Best-of-N~\cite{stiennon2022learningsummarizehumanfeedback}
      & 78.0 & 0.77 & 12.3 & 87.2 & 0.87 & 20.4 & 40.8 & 74.1 & 51.1 & 72.7 \\
      Self-Consistency~\cite{wang2023selfconsistencyimproveschainthought}
      & \underline{78.6} & 0.78 & 12.3 & 87.0 & 0.87 & 19.7 & \underline{48.8} & 78.4 & \underline{57.6} & 74.1 \\
      Univ.\ Self-Consistency~\cite{chen2023universalselfconsistencylargelanguage}
      & 69.2 & 0.69 & 11.6 & 70.8 & 0.69 & 18.8 & 19.5 & 26.8 & 20.1 & 20.8 \\
      Self-Refine~\cite{madaan2023selfrefineiterativerefinementselffeedback}
      & 62.5 & 0.61 & 29.9 & \underline{94.0} & \underline{0.94} & 18.9 & 46.8 & 106.4 & 55.1 & 95.6 \\
      Self-Eval Beam Search~\cite{xie2023selfevaluationguidedbeamsearch}
      & 60.1 & 0.58 & \textbf{1.6} & 67.6 & 0.65 & \textbf{1.6} & 18.5 & \textbf{4.3} & 22.5 & \textbf{4.3} \\
      Faithful CoT~\cite{lyu2023faithfulchainofthoughtreasoning}
      & \underline{78.6} & \underline{0.79} & 7.7 & 85.8 & 0.86 & 14.2 & 39.2 & 24.3 & 46.9 & 22.3 \\
      Tree-of-Thought~\cite{yao2023treethoughtsdeliberateproblem}
      & 72.0 & 0.71 & 16.4 & 79.4 & 0.79 & 20.6 & 37.0 & 75.0 & 47.7 & 70.2 \\
      RAP~\cite{hao2023reasoninglanguagemodelplanning}
      & 72.8 & 0.72 & 17.1 & 89.6 & 0.89 & 20.6 & 36.8 & 77.7 & 49.1 & 74.5 \\
      PNS-Optimization~\cite{yu2025causalsufficiencynecessityimproves}
      & 75.4 & 0.74 & 10.2 & 76.4 & 0.75 & 13.9 & 36.6 & 61.5 & 47.7 & 59.9 \\
      Self-Aggregation~\cite{venkatraman2026recursiveselfaggregationunlocksdeep}
      & 76.8 & 0.76 & 15.6 & 86.6 & 0.87 & 18.1 & 32.3 & 39.0 & 43.5 & 36.7 \\
  \noalign{\vskip 1pt}\cdashline{1-11}\noalign{\vskip 1.5pt}
      \textbf{CRAFT (Ours)}
      & \textbf{83.0} & \textbf{0.83} & \underline{6.6} & \textbf{96.2} & \textbf{0.96} & \underline{7.6} & \textbf{55.5} & \underline{7.2} & \textbf{63.2} & \underline{7.3} \\
  \midrule
  \rowcolor{gray!20}\multicolumn{11}{c}{\textbf{Gemini-3.1-Flash-Lite}} \\
  \midrule
      Direct
      & 74.8 & 0.75 & -- & 62.4 & 0.59 & -- & 17.5 & -- & 18.9 & -- \\
      Raw CoT~\cite{wei2023chainofthoughtpromptingelicitsreasoning}
      & 87.5 & 0.88 & 13.0 & 61.0 & 0.55 & 16.7 & 45.1 & 27.5 & 61.6 & 26.7 \\
      Best-of-N~\cite{stiennon2022learningsummarizehumanfeedback}
      & 89.1 & \underline{0.89} & 13.3 & 60.8 & 0.56 & 17.2 & 46.1 & 24.9 & 61.2 & 27.1 \\
      Self-Consistency~\cite{wang2023selfconsistencyimproveschainthought}
      & \underline{89.9} & \textbf{0.90} & 13.3 & 61.0 & 0.55 & 17.4 & \underline{55.1} & 24.9 & 63.8 & 26.8 \\
      Univ.\ Self-Consistency~\cite{chen2023universalselfconsistencylargelanguage}
      & 80.7 & 0.81 & 12.5 & 52.8 & 0.42 & 17.2 & 24.0 & 11.7 & 29.4 & \underline{12.2} \\
      Self-Refine~\cite{madaan2023selfrefineiterativerefinementselffeedback}
      & \underline{89.9} & \textbf{0.90} & 12.9 & \underline{68.6} & \underline{0.68} & 14.8 & 54.7 & 21.0 & \underline{65.8} & 25.0 \\
      Self-Eval Beam Search~\cite{xie2023selfevaluationguidedbeamsearch}
      & 80.0 & 0.80 & \textbf{2.4} & 65.8 & 0.65 & \textbf{3.8} & 37.0 & \textbf{6.8} & 48.7 & \underline{12.2} \\
      Faithful CoT~\cite{lyu2023faithfulchainofthoughtreasoning}
      & 86.7 & 0.87 & \underline{6.8} & 58.2 & 0.56 & \underline{7.2} & 46.8 & 10.7 & 59.0 & 12.9 \\
      Tree-of-Thought~\cite{yao2023treethoughtsdeliberateproblem}
      & 86.7 & 0.87 & 11.2 & 60.6 & 0.54 & 14.7 & 45.3 & 23.9 & 59.4 & 27.1 \\
      RAP~\cite{hao2023reasoninglanguagemodelplanning}
      & 86.5 & 0.87 & 11.7 & 63.6 & 0.59 & 15.7 & 49.6 & 25.3 & 61.2 & 28.4 \\
      PNS-Optimization~\cite{yu2025causalsufficiencynecessityimproves}
      & \textbf{90.3} & \textbf{0.90} & 11.1 & 59.4 & 0.52 & 13.4 & 49.4 & 23.1 & 63.2 & 24.0 \\
      Self-Aggregation~\cite{venkatraman2026recursiveselfaggregationunlocksdeep}
      & 86.3 & 0.87 & 11.5 & 64.0 & 0.60 & 13.9 & 49.0 & 18.2 & 61.8 & 20.9 \\
  \noalign{\vskip 1pt}\cdashline{1-11}\noalign{\vskip 1.5pt}
      \textbf{CRAFT (Ours)}
      & 89.3 & \underline{0.89} & 8.1 & \textbf{86.6} & \textbf{0.87} & 8.3 & \textbf{59.3} & \underline{7.8} & \textbf{73.4} & \textbf{8.1} \\
  \bottomrule
  \end{tabular}}}
  \vspace{-10pt}
  \caption{Benchmark results for two backbone models. Metrics: \emph{Acc} (\%), Macro-\emph{F1}, and \emph{Average Steps}. \textbf{Bold} = best;
  \underline{underline} = second-best within each setting. Higher is better for \emph{Acc} and \emph{F1}; lower is better for \emph{Steps}.
  F1 is omitted for mathematical reasoning datasets because each problem has a unique answer, making F1\,=\,Accuracy.}
  \label{tab:icl_main}
  \vspace{-10pt}
\end{table*}

\subsection{Experimental Results}

\paragraph{PRMBench.}
As shown in Table~\ref{tab:combined_eval} and Figure~\ref{fig:significance_forest}, providing the correct answer yields no consistent improvement across the three PRMBench dimensions. Paired Wilcoxon signed-rank tests (metrics pooled per dimension; see Appendix~\ref{app:sig_method} for details) confirm this finding: 11 of the 12 model--dimension pairs show \emph{no significant difference}. Only DeepSeek-V4-Flash on Simplicity reaches significance ($p{=}0.012$), with a negligible effect size ($d{=}0.05$). In several cases, providing correct answers even hurts performance (e.g., GPT-5.4-nano Simplicity F1 drops from 0.77 to 0.72). Detailed results are in Appendix~\ref{app:full_eval}.

\paragraph{ROSCOE.}
eSNLI and DROP show no consistent influence. CosmosQA improves for all four models ($p < 0.001$), because ROSCOE scores its traces against a field holding the correct option, which the model shows strong conformity once given the answer. On GSM8K, the \textbf{w/ Answer} setting even lowers quality for all four models ($p < 0.05$). Detailed results are in Appendix~\ref{app:full_eval}.

\paragraph{Takeaways.}
\label{Benchmark Evaluation Results}
We draw three key findings:

(1)~\textbf{In general, providing the correct answer does NOT improve reasoning ability.} Trace quality improves consistently only on CosmosQA, where the model copies the given answer and shows strong conformity to it.

(2)~\textbf{LLM reasoning flaws cannot be resolved by simply providing the correct answer.} Problems arise from the model's internal reasoning, not from uncertainty about the correct answer.

(3)~\textbf{A unified mitigation framework is necessary}, as different samples exhibit different flaw types. This motivates CRAFT below.

\subsection{Experimental Setup of CRAFT}
\label{sec:craft_setup}

As we have shown that providing the correct answer cannot improve LLMs' reasoning, we evaluate whether CRAFT's structural consensus approach succeeds. We measure label-prediction accuracy on four datasets. \textbf{Logical reasoning:} (1) FLD~\cite{morishita2024enhancingreasoningcapabilitiesllms} requires step-by-step deductive inference over natural language facts, (2) ProofWriter~\cite{tafjord2021proofwriter} asks whether a hypothesis follows from a rule base of facts and if-then rules; we use its depth-5 questions only, each of which needs a five-step deduction. \textbf{Mathematical reasoning:} (1) Omni-MATH~\cite{gao2024omnimath} and (2) OlympiadBench~\cite{he2024olympiadbenchchallengingbenchmarkpromoting} contain competition-level mathematics problems demanding advanced reasoning. We select 500 samples for each logical reasoning dataset (250 labeled "Proved" and 250 "Disproved"), and 500 samples for each mathematical reasoning dataset, with random seed = 42. For evaluation, we apply Exact-Match (EM) to the final \textsc{Proved}/\textsc{Disproved} label of each trace on the logical datasets; on the mathematical datasets, the last boxed answer is normalized and compared with the gold answer by symbolic equivalence.

We evaluate under two backbone models: \textbf{GPT-5.4-nano} (a lightweight reasoning model) and \textbf{Gemini-3.1-Flash-Lite}. On ProofWriter, and on the mathematical datasets with Gemini, a domain pass follows synthesis.


\subsection{Experimental Results}

Table~\ref{tab:icl_main} compares CRAFT against twelve baselines, details are in Appendix~\ref{app:baselines}.



\begin{figure*}[t]
  \centering
  \includegraphics[width=\linewidth]{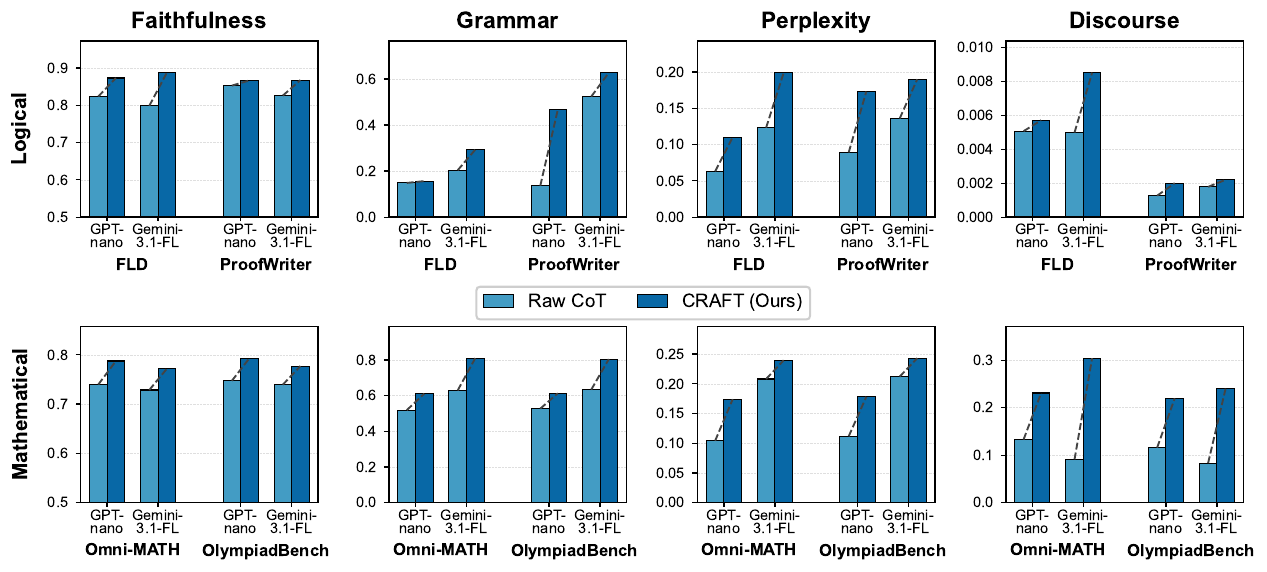}
  \vspace{-20pt}
  \caption{\textbf{ROSCOE} evaluation comparing the quality of CRAFT's post-processed traces against Raw CoT. GPT-nano: GPT-5.4-nano; Gemini-3.1-FL: Gemini-3.1-Flash-Lite. Higher is better for all metrics.}

  \label{fig:roscoe_craft}
  \vspace{-8pt}
\end{figure*}

CRAFT achieves the best accuracy in label-prediction in seven of the eight settings. Gains are notable on harder datasets: on OlympiadBench~\cite{he2024olympiadbenchchallengingbenchmarkpromoting}, CRAFT outperforms the next-best baseline by up to $7.6\%$ in accuracy while using over 65\% fewer steps. The reduction in steps comes without any accuracy trade-off: CRAFT uses fewer steps than every multi-sample baseline, because Modules~I and~II remove both \emph{Step Internal Flaws} and \emph{Step-wise Flaws}, while Module~III synthesizes one high-quality reasoning trace.



We summarize the key findings as takeaways:

\noindent$\triangleright$ \textbf{Consensus-driven trace synthesis offers higher accuracy with less steps.} CRAFT synthesizes one trace from the consensus of $K$ candidate traces. It achieves the best accuracy in seven of the eight settings across logical and mathematical reasoning datasets with less steps.


\noindent$\triangleright$ \textbf{LLMs get the majority of steps correct, even on hard problems.} While accuracy varies across  datasets, accuracy gains of CRAFT generalize across both hard and easy datasets, implying that the majority of steps are correct. Otherwise, consensus aggregation across $K$ traces would amplify errors, leading to accuracy drops.

\subsection{Quality Evaluation of Post-Processed Reasoning Traces}
\label{sec:roscoe}

We evaluate whether CRAFT improves reasoning trace quality by running ROSCOE~\cite{golovneva2023roscoesuitemetricsscoring} on CRAFT's traces and on each backbone model's raw CoT traces.

Results of ROSCOE are in Figure~\ref{fig:roscoe_craft}, which reports \emph{Faithfulness}, \emph{Grammar}, \emph{Perplexity}, and \emph{Discourse}; CRAFT is ahead on all four in all eight settings. \emph{Faithfulness} measures how well each step aligns with the problem, so CRAFT's steps stay grounded in the given facts and conditions. \emph{Grammar} scores the least grammatical step of a trace and \emph{Perplexity} the fluency of the whole chain, which indicates post-processed traces of CRAFT read better and contain no badly formed step. \emph{Discourse} is evaluated via an NLI model to check whether any step contradicts the problem, and performance gain shows that CRAFT's steps stay more consistent with the premises.

 \begin{figure}[t]
  \centering
  \includegraphics[width=\columnwidth]{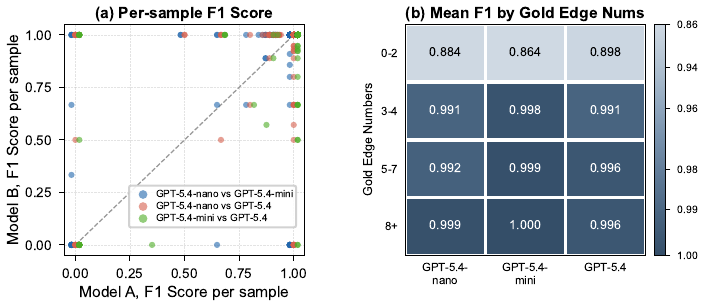}
  \vspace{-20pt}
\caption{Edge extraction F1 of $G_S$ in graph construction: per-sample F1 for each pair of models (left), and mean F1 grouped by the number of gold edges (right).}
  \label{fig:rkg robustness}
\end{figure}

\begin{table*}[t]
\centering
\setlength{\tabcolsep}{4pt}
\renewcommand{\arraystretch}{1.12}
\resizebox{\textwidth}{!}{%
{\small
\begin{tabular}{l cccc cccc}
\toprule
& \multicolumn{4}{c}{\textbf{GPT-5.4-nano}} & \multicolumn{4}{c}{\textbf{Gemini-3.1-Flash-Lite}} \\
\cmidrule(lr){2-5}\cmidrule(lr){6-9}
\textbf{Setting} & FLD & ProofWriter & Omni-MATH & Olympiad & FLD & ProofWriter & Omni-MATH & Olympiad \\
\midrule
\textbf{CRAFT} & \textbf{82.8} & \textbf{97.2} & \textbf{54.9} & \textbf{60.6} & \textbf{89.5} & \textbf{87.4} & \textbf{62.6} & \textbf{75.6} \\
w/o CRAFT & 66.1 {\scriptsize($-$16.7)} & 72.0 {\scriptsize($-$25.2)} & 6.9 {\scriptsize($-$47.6)} & 4.0 {\scriptsize($-$56.6)} & 74.8 {\scriptsize($-$14.7)} & 62.4 {\scriptsize($-$25.0)} & 17.5 {\scriptsize($-$43.7)} & 18.9 {\scriptsize($-$56.8)} \\
Single Rollout ($K$=1) & 78.4 {\scriptsize($-$4.4)} & 97.8 {\scriptsize(+0.6)} & 48.8 {\scriptsize($-$6.1)} & 58.1 {\scriptsize($-$2.4)} & 87.1 {\scriptsize($-$2.4)} & 85.4 {\scriptsize($-$2.0)} & 61.8 {\scriptsize($-$0.8)} & 73.8 {\scriptsize($-$1.8)} \\
w/o RKG & 79.4 {\scriptsize($-$3.4)} & 91.0 {\scriptsize($-$6.2)} & 45.3 {\scriptsize($-$9.5)} & 55.3 {\scriptsize($-$5.2)} & 86.8 {\scriptsize($-$2.0)} & 83.6 {\scriptsize($-$3.8)} & 62.2 {\scriptsize($-$0.4)} & 72.4 {\scriptsize($-$3.2)} \\
\noalign{\vskip 1pt}\hdashline\noalign{\vskip 1.5pt}
Replacing Jaccard with Embedding & 81.4 {\scriptsize($-$1.4)} & 98.4 {\scriptsize(+1.2)} & 55.3 {\scriptsize(+0.3)} & 60.7 {\scriptsize(+0.5)} & 89.5 {\scriptsize($\pm$0.0)} & 83.6 {\scriptsize($-$3.8)} & 61.8 {\scriptsize($-$0.8)} & 75.9 {\scriptsize(+0.2)} \\
\bottomrule
\end{tabular}}%
}
\vspace{-10pt}
\caption{Ablation study of CRAFT. Accuracy (\%), with the change compared to CRAFT with all components.}
\label{tab:ablation}
\end{table*}

\section{Noise Analysis on Graph Construction}
Graph construction $G_S$ relies on an LLM prompt to identify step dependencies. To test its robustness, we use 500 FLD problems, each annotated with a proof that lists the premises every step is derived from. We write each proof as a numbered reasoning trace and take these dependencies as gold edges. Three backbone models (GPT-5.4-nano, GPT-5.4-mini, GPT-5.4) then build $G_S$ from the same traces with identical prompts, and we score the extracted edges against the gold edges with F1. F1 varies by only about 1\% across models. In the left panel of Figure~\ref{fig:rkg robustness}, the per-sample F1 of any two models mostly coincides at 1.0; on the right, the three models reach similar mean F1 on $G_S$ at every gold edge count, so they construct graphs equally well. This confirms that graph construction does not introduce explicit model-specific noise.



\section{Ablation Study}
\label{sec:ablation}
Table~\ref{tab:ablation} validates the effectiveness of each component by removing them on all four datasets and reporting accuracy changes compared to full CRAFT framework.

Ablating the whole CRAFT pipeline leads to accuracy drops of 14.7\% or more, and removing RKG costs accuracy in all eight settings ($-$0.4\% to $-$9.5\%). Single Rollout with $K{=}1$ lowers accuracy in seven of the eight settings, by up to 6.1\%. Replacing Jaccard similarity with the cosine similarity of \texttt{all-mpnet-base-v2} embeddings lowers accuracy by up to 3.8\%. Even though there are scenarios that embedding models perform better, we still consider Jaccard is a better choice: maintaining high accuracy without additional cost on deploying embedding models.


\section{Conclusion}
 In this work, we show that providing correct answers does not consistently improve LLMs' reasoning ability, which clarifies the problem: reasoning flaws lie in the structure of generated traces, rather than in the exposure to the correct answer. We categorize reasoning flaws into \emph{Step Internal Flaws} and \emph{Step-wise Flaws}, and propose CRAFT, a unified framework that mitigates both flaw types through cross-trace consensus. CRAFT builds a consensus Reasoning Knowledge Graph (RKG), then synthesizes a high-quality trace via topological generation. CRAFT consistently outperforms baselines in label-prediction accuracy across logical and mathematical reasoning datasets. Post-processed reasoning traces achieve higher quality on fine-grained benchmark evaluation. Our work suggests that applying trace-wise consensus can be a promising direction for future LLM reasoning development.

\section*{Limitations}
Our work has the following limitations:

(1)~\textbf{Sequential CoT assumption.} CRAFT is designed for sequential Chain-of-Thought traces and may fall short on parallel or tree-structured reasoning paradigms.

(2)~\textbf{Consensus $\neq$ correctness.} The framework assumes that steps and edges appearing frequently across $K$ traces are more likely correct. On hard problems where many candidates share the same wrong reasoning pattern (e.g., a common post-hoc rationalisation), consensus can amplify rather than suppress errors. Increasing $K$ mitigates this, but does not eliminate it.

(3)~\textbf{TF-IRF on small $K$.} TF is computed over only $K$ traces per sample; with small $K$ the frequency signal may lack discriminative power, making term importance estimates noisier.

\section*{Ethical Considerations}
Our work focuses on both logical and mathematical reasoning domains. We do not address sensitive topics like race or gender. All evaluations are carried out on established benchmarks and there is no explicit ethical considerations.

\section*{Acknowledgment}
We thank \textbf{Joohyung Yun} of POSTECH for the detailed guidance on framing the paper, and \textbf{Guoxia Zhang} of Bosch (China) Investment Ltd. for valuable discussions and insights. This work was supported by the National Natural Science Foundation of China (Grant No.62506318); Guangdong Provincial Department of Education Project (Grant No.2024KQNCX028); Scientific Research Projects for the Higher-educational Institutions (Grant No.2024312096), Education Bureau of Guangzhou Municipality; Guangzhou-HKUST(GZ) Joint Funding Program (Grant No.2025A03J3957), Education Bureau of Guangzhou Municipality.

\bibliography{citations}

\clearpage
\appendix

\section{Full Pilot Study Results}
\label{app:full_eval}

Tables~\ref{tab:prmbench_full} and~\ref{tab:roscoe_full} give every metric behind Table~\ref{tab:combined_eval}. Each PRMBench row is 200 items; each ROSCOE row is 125 problems. The two settings of a row are scored on the same items, so the difference between them is paired.

The PRMBench Total rows are item-weighted means over the three dimensions, so a dimension that lost an item or two to parsing weighs slightly less. The repetition, discourse, coherence and perplexity metrics of ROSCOE stay near zero for every model and setting, so a difference of 0.01 there is large relative to their spread; we print them to three decimals. On CosmosQA the correct option is also the \texttt{hypothesis} that ROSCOE aligns each step against, so the higher faithfulness and informativeness of the w/ Answer rows there mostly reflect the model copying the option it was given.

\begin{table*}[!ht]
\centering
\caption{PRMBench, all metrics. \emph{StepAcc}: accuracy over all steps; \emph{CorrAcc}/\emph{WrongAcc}: accuracy on the correct and on the erroneous steps; \emph{1stErr}: whether the first erroneous step is located; \emph{Prec}/\emph{Recall}/\emph{F1}: correct steps as the positive class; \emph{neg-F1}: erroneous steps as the positive class. 200 items per dimension (o4-mini scored 199 on Sensitivity, GPT-5.4-nano 198 on Soundness and 197 on Sensitivity; the rest failed to parse in one setting and are dropped from both). Higher is better throughout.}
\label{tab:prmbench_full}
\resizebox{\textwidth}{!}{%
\renewcommand{\arraystretch}{1.1}
\begin{tabular}{llc cccccccc}
\toprule
\textbf{Model} & \textbf{Setting} & \textbf{Dimension} & \textbf{StepAcc} & \textbf{CorrAcc} & \textbf{WrongAcc} & \textbf{1stErr} & \textbf{Prec} & \textbf{Recall} & \textbf{F1} & \textbf{neg-F1} \\
\midrule
\multirow{8}{*}{o4-mini}
  & \multirow{4}{*}{w/ Answer}
    & Simplicity & 0.72 & 0.77 & 0.53 & 0.65 & 0.87 & 0.77 & 0.82 & 0.43 \\
  & & Soundness & 0.85 & 0.87 & 0.70 & 0.74 & 0.95 & 0.87 & 0.91 & 0.54 \\
  & & Sensitivity & 0.85 & 0.90 & 0.47 & 0.61 & 0.93 & 0.90 & 0.91 & 0.42 \\
  & & \textbf{Total} & 0.81 & 0.85 & 0.56 & 0.67 & 0.92 & 0.85 & 0.88 & 0.46 \\
\cmidrule(lr){2-11}
  & \multirow{4}{*}{w/o Answer}
    & Simplicity & 0.74 & 0.78 & 0.54 & 0.64 & 0.87 & 0.78 & 0.83 & 0.45 \\
  & & Soundness & 0.84 & 0.86 & 0.72 & 0.77 & 0.95 & 0.86 & 0.90 & 0.53 \\
  & & Sensitivity & 0.86 & 0.91 & 0.48 & 0.59 & 0.93 & 0.91 & 0.92 & 0.44 \\
  & & \textbf{Total} & 0.81 & 0.85 & 0.58 & 0.66 & 0.92 & 0.85 & 0.88 & 0.47 \\
\midrule
\multirow{8}{*}{GPT-5.4-nano}
  & \multirow{4}{*}{w/ Answer}
    & Simplicity & 0.62 & 0.62 & 0.60 & 0.64 & 0.86 & 0.62 & 0.72 & 0.38 \\
  & & Soundness & 0.69 & 0.67 & 0.78 & 0.76 & 0.95 & 0.67 & 0.79 & 0.39 \\
  & & Sensitivity & 0.75 & 0.78 & 0.54 & 0.47 & 0.93 & 0.78 & 0.85 & 0.34 \\
  & & \textbf{Total} & 0.68 & 0.69 & 0.64 & 0.62 & 0.91 & 0.69 & 0.78 & 0.37 \\
\cmidrule(lr){2-11}
  & \multirow{4}{*}{w/o Answer}
    & Simplicity & 0.67 & 0.70 & 0.52 & 0.57 & 0.86 & 0.70 & 0.77 & 0.38 \\
  & & Soundness & 0.70 & 0.70 & 0.72 & 0.75 & 0.94 & 0.70 & 0.80 & 0.38 \\
  & & Sensitivity & 0.75 & 0.79 & 0.49 & 0.50 & 0.92 & 0.79 & 0.85 & 0.32 \\
  & & \textbf{Total} & 0.70 & 0.73 & 0.58 & 0.61 & 0.91 & 0.73 & 0.81 & 0.36 \\
\midrule
\multirow{8}{*}{DeepSeek-V4-Flash}
  & \multirow{4}{*}{w/ Answer}
    & Simplicity & 0.82 & 0.92 & 0.40 & 0.47 & 0.86 & 0.92 & 0.89 & 0.47 \\
  & & Soundness & 0.84 & 0.85 & 0.74 & 0.73 & 0.96 & 0.85 & 0.90 & 0.53 \\
  & & Sensitivity & 0.85 & 0.89 & 0.54 & 0.58 & 0.94 & 0.89 & 0.91 & 0.45 \\
  & & \textbf{Total} & 0.83 & 0.89 & 0.56 & 0.59 & 0.92 & 0.89 & 0.90 & 0.48 \\
\cmidrule(lr){2-11}
  & \multirow{4}{*}{w/o Answer}
    & Simplicity & 0.77 & 0.87 & 0.39 & 0.50 & 0.85 & 0.87 & 0.86 & 0.40 \\
  & & Soundness & 0.84 & 0.86 & 0.68 & 0.71 & 0.95 & 0.86 & 0.90 & 0.51 \\
  & & Sensitivity & 0.84 & 0.89 & 0.48 & 0.54 & 0.93 & 0.89 & 0.91 & 0.42 \\
  & & \textbf{Total} & 0.82 & 0.87 & 0.51 & 0.58 & 0.91 & 0.87 & 0.89 & 0.44 \\
\midrule
\multirow{8}{*}{Gemini-3.1-Flash-Lite}
  & \multirow{4}{*}{w/ Answer}
    & Simplicity & 0.73 & 0.76 & 0.59 & 0.69 & 0.88 & 0.76 & 0.82 & 0.46 \\
  & & Soundness & 0.84 & 0.86 & 0.73 & 0.72 & 0.96 & 0.86 & 0.91 & 0.54 \\
  & & Sensitivity & 0.84 & 0.88 & 0.50 & 0.47 & 0.93 & 0.88 & 0.91 & 0.41 \\
  & & \textbf{Total} & 0.80 & 0.83 & 0.61 & 0.63 & 0.92 & 0.83 & 0.88 & 0.47 \\
\cmidrule(lr){2-11}
  & \multirow{4}{*}{w/o Answer}
    & Simplicity & 0.71 & 0.73 & 0.65 & 0.72 & 0.89 & 0.73 & 0.80 & 0.47 \\
  & & Soundness & 0.84 & 0.85 & 0.72 & 0.71 & 0.95 & 0.85 & 0.90 & 0.52 \\
  & & Sensitivity & 0.83 & 0.87 & 0.48 & 0.48 & 0.93 & 0.87 & 0.90 & 0.39 \\
  & & \textbf{Total} & 0.79 & 0.82 & 0.61 & 0.64 & 0.92 & 0.82 & 0.87 & 0.46 \\
\bottomrule
\end{tabular}}
\end{table*}

\begin{table*}[!ht]
\centering
\caption{ROSCOE in the pilot study, all twenty metrics, 125 problems per dataset and setting. The thirteen unmarked metrics need no reference chain; the seven marked \textsuperscript{$*$} compare against the gold chain and exist only for eSNLI and GSM8K. Every metric is reported on ROSCOE's own scale, higher is better; the repetition, discourse, coherence and perplexity columns are small by construction. \emph{Faith.}: faithfulness; \emph{Info-S/C}: informativeness of steps and of the chain; \emph{Faith-WW}: word-level faithfulness; \emph{Rep-W/S}: word- and step-level repetition; \emph{Gram.}/\emph{Gram-max}: mean and worst-step grammaticality; \emph{Disc.}: discourse representation; \emph{Coh.}: step-to-step coherence; \emph{PPL-S/C/max}: step, chain and worst-step perplexity; \emph{R-Align}: reasoning alignment; \emph{Ext-Hall}: external hallucination; \emph{CS-Err}: common-sense error; \emph{Cov-S/C}: semantic coverage of steps and chain.}
\label{tab:roscoe_full}
\resizebox{\textwidth}{!}{%
\renewcommand{\arraystretch}{1.1}
{\fontsize{6}{8}\selectfont
\begin{tabular}{llc cccccccccccccccccccc}
\toprule
\textbf{Model} & \textbf{Set.} & \textbf{Dataset} & \textbf{Faith.} & \textbf{Info-S} & \textbf{Info-C} & \textbf{Faith-WW} & \textbf{Rep-W} & \textbf{Rep-S} & \textbf{Gram.} & \textbf{Gram-max} & \textbf{Disc.} & \textbf{Coh.} & \textbf{PPL-S} & \textbf{PPL-C} & \textbf{PPL-max} & \textbf{R-Align\textsuperscript{$*$}} & \textbf{Ext-Hall\textsuperscript{$*$}} & \textbf{Redund.\textsuperscript{$*$}} & \textbf{CS-Err\textsuperscript{$*$}} & \textbf{Missing\textsuperscript{$*$}} & \textbf{Cov-S\textsuperscript{$*$}} & \textbf{Cov-C\textsuperscript{$*$}} \\
\midrule
\multirow{8}{*}{o4-mini}
& \multirow{4}{*}{w/ Answer}
  & CosmosQA & 0.82 & 0.81 & 0.92 & 0.93 & 0.057 & 0.071 & 0.95 & 0.82 & 0.261 & 0.318 & 0.005 & 0.033 & 0.002 & --- & --- & --- & --- & --- & --- & --- \\
& & DROP & 0.83 & 0.81 & 0.92 & 0.94 & 0.054 & 0.069 & 0.96 & 0.86 & 0.044 & 0.288 & 0.007 & 0.067 & 0.003 & --- & --- & --- & --- & --- & --- & --- \\
& & eSNLI & 0.71 & 0.78 & 0.87 & 0.91 & 0.056 & 0.079 & 0.83 & 0.57 & 0.221 & 0.096 & 0.002 & 0.036 & 0.000 & 0.75 & 0.80 & 0.59 & 0.94 & 0.79 & 0.92 & 0.89 \\
& & GSM8K & 0.82 & 0.84 & 0.96 & 0.92 & 0.039 & 0.050 & 0.94 & 0.81 & 0.245 & 0.116 & 0.005 & 0.076 & 0.001 & 0.85 & 0.93 & 0.76 & 0.88 & 0.69 & 0.95 & 0.96 \\
\cmidrule(lr){2-23}
& \multirow{4}{*}{w/o Answer}
  & CosmosQA & 0.80 & 0.79 & 0.92 & 0.93 & 0.058 & 0.062 & 0.96 & 0.83 & 0.279 & 0.372 & 0.005 & 0.038 & 0.002 & --- & --- & --- & --- & --- & --- & --- \\
& & DROP & 0.83 & 0.81 & 0.93 & 0.94 & 0.042 & 0.055 & 0.96 & 0.85 & 0.057 & 0.280 & 0.008 & 0.072 & 0.004 & --- & --- & --- & --- & --- & --- & --- \\
& & eSNLI & 0.71 & 0.78 & 0.87 & 0.91 & 0.047 & 0.067 & 0.84 & 0.56 & 0.210 & 0.058 & 0.002 & 0.038 & 0.000 & 0.75 & 0.79 & 0.58 & 0.94 & 0.79 & 0.92 & 0.89 \\
& & GSM8K & 0.83 & 0.85 & 0.96 & 0.92 & 0.039 & 0.048 & 0.95 & 0.83 & 0.308 & 0.187 & 0.005 & 0.080 & 0.002 & 0.85 & 0.94 & 0.77 & 0.87 & 0.67 & 0.94 & 0.96 \\
\midrule
\multirow{8}{*}{GPT-5.4-nano}
& \multirow{4}{*}{w/ Answer}
  & CosmosQA & 0.83 & 0.81 & 0.91 & 0.94 & 0.049 & 0.059 & 0.91 & 0.71 & 0.302 & 0.464 & 0.005 & 0.033 & 0.002 & --- & --- & --- & --- & --- & --- & --- \\
& & DROP & 0.84 & 0.81 & 0.92 & 0.94 & 0.049 & 0.058 & 0.94 & 0.83 & 0.037 & 0.402 & 0.007 & 0.058 & 0.003 & --- & --- & --- & --- & --- & --- & --- \\
& & eSNLI & 0.69 & 0.78 & 0.86 & 0.91 & 0.049 & 0.068 & 0.84 & 0.59 & 0.224 & 0.199 & 0.002 & 0.043 & 0.000 & 0.73 & 0.80 & 0.59 & 0.94 & 0.77 & 0.91 & 0.88 \\
& & GSM8K & 0.81 & 0.82 & 0.96 & 0.91 & 0.038 & 0.057 & 0.94 & 0.81 & 0.322 & 0.169 & 0.004 & 0.072 & 0.001 & 0.85 & 0.93 & 0.77 & 0.88 & 0.71 & 0.96 & 0.96 \\
\cmidrule(lr){2-23}
& \multirow{4}{*}{w/o Answer}
  & CosmosQA & 0.82 & 0.79 & 0.92 & 0.94 & 0.049 & 0.045 & 0.91 & 0.72 & 0.260 & 0.423 & 0.005 & 0.037 & 0.003 & --- & --- & --- & --- & --- & --- & --- \\
& & DROP & 0.84 & 0.80 & 0.92 & 0.94 & 0.044 & 0.046 & 0.95 & 0.86 & 0.067 & 0.455 & 0.007 & 0.062 & 0.004 & --- & --- & --- & --- & --- & --- & --- \\
& & eSNLI & 0.69 & 0.78 & 0.86 & 0.91 & 0.049 & 0.065 & 0.84 & 0.62 & 0.240 & 0.167 & 0.002 & 0.045 & 0.000 & 0.72 & 0.78 & 0.56 & 0.94 & 0.77 & 0.90 & 0.88 \\
& & GSM8K & 0.81 & 0.83 & 0.96 & 0.91 & 0.039 & 0.057 & 0.94 & 0.81 & 0.266 & 0.175 & 0.005 & 0.072 & 0.002 & 0.85 & 0.93 & 0.76 & 0.87 & 0.68 & 0.95 & 0.96 \\
\midrule
\multirow{8}{*}{DeepSeek-V4-Flash}
& \multirow{4}{*}{w/ Answer}
  & CosmosQA & 0.84 & 0.81 & 0.91 & 0.94 & 0.051 & 0.056 & 0.92 & 0.76 & 0.325 & 0.440 & 0.009 & 0.043 & 0.004 & --- & --- & --- & --- & --- & --- & --- \\
& & DROP & 0.83 & 0.80 & 0.92 & 0.94 & 0.049 & 0.061 & 0.96 & 0.87 & 0.044 & 0.357 & 0.006 & 0.068 & 0.002 & --- & --- & --- & --- & --- & --- & --- \\
& & eSNLI & 0.69 & 0.77 & 0.86 & 0.91 & 0.050 & 0.069 & 0.85 & 0.59 & 0.201 & 0.144 & 0.003 & 0.051 & 0.000 & 0.73 & 0.79 & 0.59 & 0.94 & 0.78 & 0.90 & 0.89 \\
& & GSM8K & 0.82 & 0.84 & 0.96 & 0.92 & 0.042 & 0.051 & 0.95 & 0.85 & 0.246 & 0.075 & 0.005 & 0.080 & 0.001 & 0.86 & 0.93 & 0.76 & 0.89 & 0.72 & 0.96 & 0.97 \\
\cmidrule(lr){2-23}
& \multirow{4}{*}{w/o Answer}
  & CosmosQA & 0.82 & 0.79 & 0.92 & 0.93 & 0.048 & 0.040 & 0.94 & 0.78 & 0.346 & 0.504 & 0.010 & 0.051 & 0.005 & --- & --- & --- & --- & --- & --- & --- \\
& & DROP & 0.83 & 0.80 & 0.92 & 0.94 & 0.044 & 0.052 & 0.96 & 0.87 & 0.063 & 0.364 & 0.009 & 0.075 & 0.003 & --- & --- & --- & --- & --- & --- & --- \\
& & eSNLI & 0.68 & 0.77 & 0.86 & 0.91 & 0.048 & 0.068 & 0.85 & 0.59 & 0.210 & 0.127 & 0.003 & 0.050 & 0.000 & 0.73 & 0.79 & 0.58 & 0.94 & 0.78 & 0.90 & 0.88 \\
& & GSM8K & 0.82 & 0.84 & 0.96 & 0.92 & 0.042 & 0.051 & 0.95 & 0.86 & 0.190 & 0.122 & 0.007 & 0.081 & 0.002 & 0.86 & 0.93 & 0.76 & 0.87 & 0.68 & 0.95 & 0.97 \\
\midrule
\multirow{8}{*}{Gemini-3.1-Flash-Lite}
& \multirow{4}{*}{w/ Answer}
  & CosmosQA & 0.82 & 0.79 & 0.91 & 0.93 & 0.056 & 0.061 & 0.96 & 0.87 & 0.402 & 0.567 & 0.011 & 0.046 & 0.005 & --- & --- & --- & --- & --- & --- & --- \\
& & DROP & 0.83 & 0.80 & 0.92 & 0.94 & 0.043 & 0.056 & 0.95 & 0.85 & 0.058 & 0.364 & 0.012 & 0.081 & 0.005 & --- & --- & --- & --- & --- & --- & --- \\
& & eSNLI & 0.68 & 0.76 & 0.86 & 0.91 & 0.052 & 0.080 & 0.87 & 0.68 & 0.208 & 0.128 & 0.002 & 0.044 & 0.000 & 0.74 & 0.79 & 0.59 & 0.94 & 0.78 & 0.90 & 0.89 \\
& & GSM8K & 0.82 & 0.83 & 0.95 & 0.93 & 0.039 & 0.047 & 0.94 & 0.78 & 0.228 & 0.134 & 0.010 & 0.099 & 0.003 & 0.86 & 0.93 & 0.77 & 0.88 & 0.69 & 0.96 & 0.96 \\
\cmidrule(lr){2-23}
& \multirow{4}{*}{w/o Answer}
  & CosmosQA & 0.80 & 0.78 & 0.91 & 0.93 & 0.048 & 0.046 & 0.96 & 0.90 & 0.330 & 0.526 & 0.012 & 0.053 & 0.006 & --- & --- & --- & --- & --- & --- & --- \\
& & DROP & 0.83 & 0.80 & 0.92 & 0.94 & 0.051 & 0.063 & 0.95 & 0.85 & 0.064 & 0.406 & 0.013 & 0.082 & 0.006 & --- & --- & --- & --- & --- & --- & --- \\
& & eSNLI & 0.69 & 0.76 & 0.87 & 0.91 & 0.051 & 0.077 & 0.87 & 0.67 & 0.155 & 0.128 & 0.006 & 0.048 & 0.001 & 0.73 & 0.79 & 0.58 & 0.94 & 0.78 & 0.90 & 0.89 \\
& & GSM8K & 0.83 & 0.84 & 0.95 & 0.93 & 0.036 & 0.047 & 0.96 & 0.85 & 0.265 & 0.216 & 0.011 & 0.106 & 0.004 & 0.86 & 0.95 & 0.79 & 0.86 & 0.65 & 0.95 & 0.96 \\
\bottomrule
\end{tabular}}}
\end{table*}

\section{Significance Testing}
\label{app:sig_method}

Figure~\ref{fig:significance_forest} reports paired Wilcoxon signed-rank tests with 95\% bootstrap confidence intervals. Each panel pools several metrics into one test, as described below.

\paragraph{Pooling.}
Take one model and one PRMBench dimension, say Soundness, with $N$ items. Item $i$ has a w/ Answer score $g_i^{(m)}$ and a w/o Answer score $b_i^{(m)}$ for each metric $m$ in the set $\mathcal{M}$. We concatenate the paired differences of all metrics into one vector of length $N|\mathcal{M}|$,
\begin{multline*}
\mathbf{d} = \bigl(g_1^{(1)}\!-\!b_1^{(1)},\;\ldots,\;g_N^{(1)}\!-\!b_N^{(1)},\;\ldots,\\
g_1^{(|\mathcal{M}|)}\!-\!b_1^{(|\mathcal{M}|)},\;\ldots,\;g_N^{(|\mathcal{M}|)}\!-\!b_N^{(|\mathcal{M}|)}\bigr),
\end{multline*}
and run the test and the bootstrap on $\mathbf{d}$. For PRMBench, $\mathcal{M} = \{\text{StepAcc}, \text{1stErr}, \text{F1}\}$ and the pooling is per dimension, giving $N|\mathcal{M}| = 600$ (a little less where an item failed to parse). For ROSCOE, $\mathcal{M} = \{\text{Faithfulness}, \text{Info-Step}, \text{Info-Chain}, \text{Grammar}\}$ and the pooling is per dataset, giving 500.

\paragraph{Why pooling is safe here.}
All four ROSCOE metrics and all three PRMBench metrics are bounded in $[0,1]$ and are accuracies or alignment scores, so no metric dominates the pooled vector by scale. The Wilcoxon test is rank-based and does not change under a monotone rescaling of any one metric.

\paragraph{Statistics.}
For each pooled vector we report the mean difference with a 95\% interval from 5{,}000 bootstrap resamples, the two-sided Wilcoxon $p$-value, and a paired Cohen's $d = \bar{d}/s_d$. The markers in the figure are {\color[HTML]{E67E22}$*$} for $p<0.05$, {\color[HTML]{E67E22}$**$} for $p<0.01$ and {\color[HTML]{E67E22}$***$} for $p<0.001$.

\paragraph{What the tests say.}
On PRMBench, 11 of the 12 model--dimension tests are not significant. The exception is DeepSeek-V4-Flash on Simplicity, where the answer raises the pooled score by 0.02 ($p{=}0.012$, $d{=}0.05$); pooled over all three dimensions, GPT-5.4-nano's F1 falls from 0.756 to 0.735 ($p{=}0.034$). On ROSCOE, CosmosQA is positive for all four models ($p<0.001$, $+0.004$ to $+0.009$), GSM8K is negative for all four ($p{=}0.003$ for o4-mini, $0.030$ for GPT-5.4-nano, $<0.001$ for the other two; $-0.003$ to $-0.010$), eSNLI is not significant for any model, and DROP is significant only for DeepSeek-V4-Flash, in the negative direction ($-0.002$, $p{=}0.010$).

\section{Datasets}
\label{app:dataset_details}

We use eight datasets in two parts. PRMBench and the four ROSCOE datasets form the pilot study of Section~\ref{sec:label_guidance_results}. FLD, ProofWriter, Omni-MATH and OlympiadBench are the benchmark of Section~\ref{sec:experiments}, on which CRAFT is compared with the baselines and on which ROSCOE is run a second time to score the traces.

\subsection*{Pilot study}

\noindent\textbf{PRMBench}~\cite{song2025prmbenchfinegrainedchallengingbenchmark} asks a model to act as a process reward model: given a mathematical solution into which an error has been injected, mark each step valid or invalid. Its nine error types fall into three dimensions, and we sample 200 items from each: \emph{Simplicity} (redundant and circular steps), \emph{Soundness} (counterfactual steps, contradictions between steps, domain inconsistency, and over-confident claims), and \emph{Sensitivity} (missing conditions, deception, and problems with several valid solutions). The model returns a validity score and a redundancy score per step as JSON, in the output format of PRMBench's own critics, and PRMBench's scorer turns them into step accuracy, first-error localization and F1. In the w/ Answer setting the prompt carries one extra line, the correct final answer taken from the unmodified solution; the steps scored are the modified ones in both settings.

\noindent\textbf{ROSCOE}~\cite{golovneva2023roscoesuitemetricsscoring} is a suite of reference-free and reference-based metrics for reasoning chains, packaged with four datasets that we take 125 problems from each: CosmosQA and DROP (reading comprehension), eSNLI (natural language inference) and GSM8K (arithmetic word problems). The model writes a numbered chain for each problem and ROSCOE scores the chain. The seven reference-based metrics need a gold chain, which only eSNLI and GSM8K supply.

\subsection*{Benchmark}

We use two logical and two mathematical datasets with 500 problems each, chosen so that no backbone is near the ceiling before any method is applied. The Direct and Raw CoT rows of Table~\ref{tab:icl_main} show how much room each backbone leaves on each dataset.

\noindent\textbf{FLD}~\cite{morishita2024enhancingreasoningcapabilitiesllms} is generated from propositional and first-order templates, so its facts are nonsense sentences and a model cannot answer from world knowledge. Each problem is a set of facts and a hypothesis with a binary label, \textsc{Proved} or \textsc{Disproved}. We take 250 of each.

\noindent\textbf{ProofWriter}~\cite{tafjord2021proofwriter} is a rule base of facts and if-then rules with a hypothesis to decide. We use question depth 5, where the answer needs a five-step deduction, and the RelNeg-OWA configuration, which has relational predicates with negation under the open-world assumption; it is the configuration on which both backbones had the most room to improve in our preliminary runs. Problems come from the test and validation splits only, and we keep the binary \textsc{Proved}/\textsc{Disproved} questions, 250 of each. The \textsc{Unknown} label exists in ProofWriter and in FLD; we leave it out, as the earlier FOLIO protocol did, because a framework built on consensus among traces should be judged on whether it reaches the right decision, and an undecidable class would mix its failures with real undecidability.

\noindent\textbf{Omni-MATH}~\cite{gao2024omnimath} is a collection of olympiad problems with reference answers. We drop the problems that cannot be scored automatically instead of counting them as wrong: those whose statement depends on a figure, and those whose gold answer describes a family of solutions (``all positive integers $n$ whose prime factors are $1 \bmod 4$'') instead of a value. Keeping them would cost every method the same points and would test the answer reader more than the methods.

\noindent\textbf{OlympiadBench}~\cite{he2024olympiadbenchchallengingbenchmarkpromoting} contains competition problems from national and international olympiads. We use the text-only English competition subset with a final answer (\texttt{OE\_TO\_maths\_en\_COMP}, 674 problems), stratified to 125 problems from each of algebra, combinatorics, geometry and number theory. Its theorem-proving problems have no final answer to read and are not used.

\noindent\textbf{What was dropped.}
Our preliminary runs used FOLIO~\cite{han2024folionaturallanguagereasoning} and GSM8K~\cite{cobbe2021trainingverifierssolvemath} as well. Both backbones were already close to the ceiling on them without any method, so a method could gain at most a point or two and methods could not be told apart. We replaced them with ProofWriter at depth 5 and Omni-MATH. Both mathematics datasets have a single answer per problem, so accuracy and F1 would be the same number and F1 is left out for them.

\section{Baselines and Computational Cost}
\label{app:baselines}

CRAFT is compared with twelve training-free settings run on the same backbone, the same 500 problems and the same reader. Every multi-sample baseline draws 10 candidates. CRAFT draws $K{=}5$ and spends further calls on RKG construction and step-wise synthesis, about 18 per problem in all (Table~\ref{tab:compute}), so 10 candidates give the baselines a comparable budget. RAP, PNS-Optimization and the tree-search Tree-of-Thought are reimplemented from their papers' descriptions rather than run from the authors' code, so their numbers are comparable with the other rows of Table~\ref{tab:icl_main} and not with the originals.

\paragraph{Direct} asks for the label with no reasoning. It is the floor a backbone reaches from the problem alone, and the ``w/o CRAFT'' row of the ablation.

\paragraph{Raw CoT}~\cite{wei2023chainofthoughtpromptingelicitsreasoning} asks the model to reason step by step in one call. Its response is also the raw side of the trace-quality comparison in Section~\ref{sec:roscoe}.

\paragraph{Best-of-N}~\cite{stiennon2022learningsummarizehumanfeedback} draws $N{=}10$ traces, has the model score each of its own traces, and returns the highest scored one.

\paragraph{Self-Consistency}~\cite{wang2023selfconsistencyimproveschainthought} draws 10 traces and returns the majority answer.

\paragraph{Universal Self-Consistency}~\cite{chen2023universalselfconsistencylargelanguage} replaces the hard vote with one more call in which the model reads the ten candidate answers and picks the most consistent. It avoids format-sensitive answer extraction at the cost of one call.

\paragraph{Self-Refine}~\cite{madaan2023selfrefineiterativerefinementselffeedback} revises a single trace through ten rounds of self-critique and rewrite.

\paragraph{Self-Eval Beam Search}~\cite{xie2023selfevaluationguidedbeamsearch} grows a trace step by step and keeps a beam of width 2 ranked by the model's own step scores, as in the original paper. The beam prunes aggressively, so its traces are among the shortest in Table~\ref{tab:icl_main}, and on the mathematics datasets it loses most of the accuracy along with the steps.

\paragraph{Faithful CoT}~\cite{lyu2023faithfulchainofthoughtreasoning} has the model translate the problem into a symbolic program and executes the program for the answer. Its traces are shorter than CRAFT's on FLD with both backbones and on ProofWriter with Gemini.

\paragraph{Tree-of-Thought}~\cite{yao2023treethoughtsdeliberateproblem} searches a tree of partial reasoning states with branching width 5 and depth 2, scoring each state with the model, with the original paper's parameters and no per-task tuning.

\paragraph{RAP}~\cite{hao2023reasoninglanguagemodelplanning} treats reasoning as planning with the model as a world model and runs Monte Carlo tree search with UCB selection for 8 iterations.

\paragraph{PNS-Optimization}~\cite{yu2025causalsufficiencynecessityimproves} scores every step of 10 candidate chains by its probability of necessity and sufficiency for the final answer and keeps the steps above the paper's default threshold of 0.5.

\paragraph{Self-Aggregation}~\cite{venkatraman2026recursiveselfaggregationunlocksdeep} recursively recombines 10 candidate traces into one refined answer over several rounds of summarization, rather than selecting one candidate.

\begin{table*}[t]
\centering
\setlength{\tabcolsep}{8pt}
\renewcommand{\arraystretch}{1.15}
{\small
\begin{tabular}{clcc}
\toprule
\textbf{Module} & \textbf{Operation} & \textbf{Calls} & \textbf{Model} \\
\midrule
\multirow{3}{*}{I}  & Generate $K$ traces        & $K$       & backbone \\
                    & TF-IRF term extraction     & 0         & --- \\
                    & Z-score step filter        & 0         & --- \\
\midrule
\multirow{2}{*}{II} & Build per-trace RKGs       & $K$       & backbone \\
                    & Consensus RKG, edge and node filter & 0   & --- \\
\midrule
III & Topology-guided synthesis, one call per node & $n^{*}$ & backbone \\
\midrule
--- & Domain pass (ProofWriter; mathematics with Gemini) & a few & backbone \\
\midrule
    & \textbf{Total per problem}  & \multicolumn{2}{c}{$2K + n^{*}$ (+ domain pass)} \\
\bottomrule
\end{tabular}}
\vspace{-6pt}
\caption{LLM calls per problem. $K{=}5$; $n^{*}$ is the number of nodes in the consensus RKG, 8--9 on average. The total is about 18 calls. Every call goes to the column's backbone, GPT-5.4-nano or Gemini-3.1-Flash-Lite.}
\vspace{-6pt}
\label{tab:compute}
\end{table*}

\noindent\textbf{Cost.} TF-IRF extraction, z-score filtering and graph filtering are computation without LLM calls. The calls are the $K$ rollouts, the $K$ graph extractions, and one call per consensus node in synthesis, about 18 per problem at $K{=}5$ and 8 nodes. Self-Consistency and Best-of-N with 10 candidates make 10 to 11 calls, Self-Refine one draft and ten rounds of critique and rewrite, Self-Eval Beam Search about $K \times B$ for beam width $B$, and RAP and Tree-of-Thought more still, since each expands several branches per step. Compared with simple voting, CRAFT's extra calls go to graph extraction and step-wise synthesis.

\section{Evaluation Details}
\label{app:eval_details}

\paragraph{Reading the label.}
One reader scores every method. For FLD and ProofWriter it takes the last \texttt{\_\_PROVED\_\_} or \texttt{\_\_DISPROVED\_\_} token in the trace. For Omni-MATH and OlympiadBench it takes the last \verb|\boxed{}| value, normalizes it (fractions, radicals, units, ordering inside tuples and sets), and compares it with the gold answer by symbolic equivalence through \texttt{sympy}, falling back to numeric comparison at the gold answer's precision. A trace with no readable answer counts as wrong. There is no LLM judge anywhere in the scoring.

\paragraph{F1 and accuracy.}
F1 is the macro average over \textsc{Proved} and \textsc{Disproved}, which differs from accuracy when a method predicts one label more often than the other; Universal Self-Consistency and Self-Eval Beam Search on ProofWriter with Gemini are the clearest cases in Table~\ref{tab:icl_main}. The mathematics datasets have one answer per problem and no classes to average, so they report accuracy alone.

\paragraph{Counting steps.}
A step is a line that declares itself one, ``Step 3: \ldots'', plus the conclusion line when it is not already among the steps; a trace that declares no steps counts each non-empty line as one. Nothing is capped. The same function counts CRAFT's traces and every baseline's, so the Steps column measures every method the same way.

\paragraph{Dropped problems.}
A few problems could not be scored for any method. The API gateway refused some mathematics prompts outright (a content filter on tokens such as \texttt{gcd}) and returned nothing for four FLD problems, so they have no traces. An empty consensus RKG does not drop a problem: Module~III then writes the voted verdict on the logical datasets and a term-guided trace on the mathematical ones. The one exception is an FLD problem with GPT-5.4-nano whose traces reach no verdict. These leave 492 to 497 problems in the mathematics cells, 495 and 496 on FLD, and 500 on ProofWriter. Every method in a column is scored on the same problems.

\paragraph{Trace quality.}
ROSCOE scores whatever chain it is given, so it is run on the same traces that produce Table~\ref{tab:icl_main}: the Raw CoT response from the baseline run is the raw side and CRAFT's synthesized trace the other, both after the same light normalization (markdown removed, mathematical notation unified). A handful of raw chains on the Gemini mathematics cells exceeded 8{,}192 tokens or did not finish scoring inside the compute budget and are absent from the raw side. FineLogic and ReCEval~\cite{prasad2023recevalevaluatingreasoningchains} were run in our preliminary runs only; the benchmark traces are scored with ROSCOE.

\section{Domain Passes after Synthesis}
\label{app:domain_passes}

Two datasets get a further pass after Module~III. Both use the column's backbone and both are part of the pipeline whose accuracy Table~\ref{tab:icl_main} reports.

\paragraph{ProofWriter: closed-world recheck.}
ProofWriter's rules can prove a hypothesis only from what the rule base states, and a synthesized trace sometimes concludes \textsc{Disproved} for a hypothesis that is provable, or the reverse. A second pass asks the backbone to attempt a proof of the hypothesis from the rule base under the closed-world assumption; when the attempt contradicts the synthesized conclusion, the proof search is written into the trace as numbered steps and the conclusion is resolved to match. The pass runs in two directions, first attempting to prove and then to resolve the remaining disagreements. With GPT-5.4-nano the resulting trace is then restated by one rewrite call with the conclusion pinned, so that the restatement cannot change the label.

\paragraph{Mathematics with Gemini: adjudication.}
When the $K$ traces of a problem split between two answers, a single call asks the backbone to adjudicate between them from the problem statement, and the derivation it gives is written back into the trace as steps rather than appended as a note. The trace then opens with the problem's own statement of what is asked. GPT-5.4-nano's mathematics cells do not use this pass.

\section{Hyperparameters}
\label{app:hyperparams}

Table~\ref{tab:hyperparams} lists the values used for Tables~\ref{tab:icl_main} and~\ref{tab:ablation}. They were set once and not tuned per dataset or per backbone.

\paragraph{How the values were chosen.}
$K{=}5$ balances diversity against cost. The step frequency threshold $\beta$ and the edge filtering threshold $\theta$ are both $0.3$. A consensus term must appear in at least 30\% of the steps. An edge must reach $W(e)\ge 0.3$, that is, its weights $W_S(e)$ summed over the traces that contain it must reach $0.3K = 1.5$. Since $W_S(e)\le 1$, an edge found in only one trace never passes, and an edge with $W_S(e)=0.7$ in every trace that contains it needs three of the five. The z-score cutoff $\gamma{=}{-1.0}$ removes a step whose term overlap with the consensus terms is more than one standard deviation below its group's mean, a middle setting between $-0.5$ and $-1.5$. In the edge weight $W_S(e) = 0.7\,\mathrm{conf}(e) + 0.3\,\mathrm{Jaccard}(u,v)$, $\mathrm{conf}(e)$ is the confidence the extractor reports for that dependency, and it carries most of the weight because Jaccard alone cannot recognize two phrasings of one inference. $\alpha$ is the TF-IRF threshold of $T_{\text{Con}}$; a term below it is also not shown to Module~III as a key term.

\section{Filtering Statistics}
\label{app:filter_stats}

Table~\ref{tab:zscore_stats} counts the steps that each pass of the filter removed on the eight runs that Table~\ref{tab:icl_main} reports. The first pass, the z-score filter of Module~I against the TF-IRF consensus terms, removes 4.0--6.1\% of the steps on the two logical datasets and 7.7--11.2\% on the two mathematical ones. The second pass, the edge and node filtering of Module~II applied to each trace's steps once the consensus RKG is built, removes a further 35--53\% on the logical datasets and 45--70\% on the mathematical ones. A step survives the second pass only when the dependency it builds on reaches $W(e)\ge\theta$, which with $K{=}5$ needs two or three traces to draw it; a step that only one trace takes does not survive. GPT-5.4-nano loses more than Gemini on the logical datasets (53\% against 35\% on FLD) because its five traces agree less often on how to decompose the same proof.

\begin{table}[!t]
\centering
\setlength{\tabcolsep}{3pt}
\renewcommand{\arraystretch}{1.1}
\resizebox{\linewidth}{!}{%
{\small
\begin{tabular}{l r rr rr}
\toprule
& & \multicolumn{2}{c}{\textbf{Pass 1 (z-score)}} & \multicolumn{2}{c}{\textbf{Pass 2 (RKG)}} \\
\cmidrule(lr){3-4}\cmidrule(lr){5-6}
\textbf{Dataset} & \textbf{Steps} & \textbf{Del} & \textbf{\%} & \textbf{Del} & \textbf{\%} \\
\midrule
FLD (nano) & 20,926 & 847 & 4.0 & 10,717 & 53.4 \\
ProofWriter (nano) & 22,187 & 1,219 & 5.5 & 9,483 & 45.2 \\
FLD (Gemini) & 21,966 & 1,340 & 6.1 & 7,138 & 34.6 \\
ProofWriter (Gemini) & 22,165 & 1,246 & 5.6 & 7,538 & 36.0 \\
\midrule
Omni-MATH (nano) & 17,823 & 1,365 & 7.7 & 11,481 & 69.8 \\
Olympiad (nano) & 18,400 & 1,608 & 8.7 & 11,749 & 70.0 \\
Omni-MATH (Gemini) & 19,079 & 2,095 & 11.0 & 8,251 & 48.6 \\
Olympiad (Gemini) & 20,070 & 2,249 & 11.2 & 7,950 & 44.6 \\
\bottomrule
\end{tabular}}}
\vspace{-6pt}
\caption{Steps removed by each filtering pass on the eight reported runs ($K{=}5$, 500 problems each). \emph{Steps}: reasoning steps across all $K$ traces before filtering. Pass~1 is the z-score filter of Module~I; Pass~2 is the edge and node filtering of Module~II, applied to each trace's steps once the consensus RKG is built. Pass~1 percentages are of the original step count, Pass~2 percentages of the steps Pass~1 left. The last step of each trace is never removed.}
\vspace{-6pt}
\label{tab:zscore_stats}
\end{table}

Table~\ref{tab:rkg_stats} says which filter caught a step in the second pass. Edge filtering (the step's incoming edges have $W(e)$ below $\theta$) accounts for 94--99\% of the flagged steps on every dataset, and node filtering (an isolated node with neither in- nor out-edges) for the rest, 1--6\%. On the mathematics datasets a third check, for steps whose arithmetic does not hold, catches under 1\%. A step that fails more than one check is counted under the first, so the columns add up to the total. The total counts flagged steps and exceeds the removals of Table~\ref{tab:zscore_stats} because a trace's last step can be flagged but is always kept.

\begin{table}[!t]
\centering
\setlength{\tabcolsep}{5pt}
\renewcommand{\arraystretch}{1.1}
\resizebox{\linewidth}{!}{%
\begin{tabular}{l r r r r}
\toprule
\textbf{Dataset} & \textbf{Removed} & \textbf{Edges} & \textbf{Nodes} & \textbf{Math check} \\
\midrule
FLD (nano) & 12,947 & 12,277 (95\%) & 670 (5\%) & --- \\
ProofWriter (nano) & 11,201 & 10,612 (95\%) & 589 (5\%) & --- \\
FLD (Gemini) & 8,946 & 8,673 (97\%) & 273 (3\%) & --- \\
ProofWriter (Gemini) & 9,352 & 8,756 (94\%) & 596 (6\%) & --- \\
\midrule
Omni-MATH (nano) & 13,687 & 13,487 (99\%) & 200 (1\%) & 0 \\
Olympiad (nano) & 13,912 & 13,667 (98\%) & 241 (2\%) & 4 ($<$1\%) \\
Omni-MATH (Gemini) & 9,915 & 9,760 (98\%) & 139 (1\%) & 16 ($<$1\%) \\
Olympiad (Gemini) & 9,566 & 9,305 (97\%) & 228 (2\%) & 33 ($<$1\%) \\
\bottomrule
\end{tabular}%
}
\vspace{-6pt}
\caption{What Pass~2 removed, by filter. \emph{Edges}: steps whose incoming edges have $W(e)$ below $\theta$; \emph{Nodes}: isolated nodes with neither in- nor out-edges; \emph{Math check}: steps whose arithmetic the mathematics-domain check found false. A step caught by more than one filter is counted under the first; \emph{Removed} counts flagged steps, including last steps, which are always kept.}
\vspace{-6pt}
\label{tab:rkg_stats}
\end{table}

\begin{table}[b]
\centering
\setlength{\tabcolsep}{3pt}
\renewcommand{\arraystretch}{1.1}
\resizebox{\linewidth}{!}{%
{\small
\begin{tabular}{ll l rr r}
\toprule
\textbf{Backbone} & \textbf{Dataset} & \textbf{Strongest baseline} & $b$ & $c$ & $p$ \\
\midrule
GPT-5.4-nano & FLD & Faithful CoT & 54 & 33 & 0.031 \\
 & ProofWriter & Self-Refine & 26 & 15 & 0.117 \\
 & Omni-MATH & Self-Consistency & 55 & 27 & 0.003 \\
 & OlympiadBench & Self-Consistency & 61 & 33 & 0.005 \\
\midrule
Gemini-3.1-Flash-Lite & FLD & PNS-Optimization & 22 & 27 & 0.568 \\
 & ProofWriter & Self-Refine & 118 & 28 & $<$0.001 \\
 & Omni-MATH & Self-Consistency & 51 & 36 & 0.133 \\
 & OlympiadBench & Self-Refine & 75 & 37 & $<$0.001 \\
\bottomrule
\end{tabular}}}
\vspace{-6pt}
\caption{Exact two-sided McNemar tests of CRAFT against the most accurate baseline of each cell, on the problems both answered. $b$: problems CRAFT gets right and the baseline wrong; $c$: the reverse.}
\vspace{-6pt}
\label{tab:mcnemar}
\end{table}

\section{Confidence Intervals and Paired Tests}
\label{app:ci}

Table~\ref{tab:ci} gives the 95\% Wilson score interval for every accuracy in Table~\ref{tab:icl_main}, computed from the $n$ of that cell. At $n{\approx}500$ the interval is about 8 points wide near 50\% accuracy and about 3.5 points wide near 95\%, so many of the differences among baselines in one column are within each other's intervals, and a difference of a point or two between CRAFT and the strongest baseline is not one the intervals can resolve.

The paired test is stronger, because it uses which problems each method got right rather than two marginal totals. Table~\ref{tab:mcnemar} reports an exact two-sided McNemar test between CRAFT and the most accurate baseline of each cell, on the problems both answered. CRAFT is ahead in seven of the eight cells, and its lead is significant at $p<0.05$ in five: FLD, Omni-MATH and OlympiadBench with GPT-5.4-nano, and ProofWriter and OlympiadBench with Gemini. It is not significant on ProofWriter with GPT-5.4-nano ($p{=}0.12$) or on Omni-MATH with Gemini ($p{=}0.13$), where CRAFT is ahead by 2.2 and 3.0 points on the problems both answered; on FLD with Gemini it is 1.0 point behind PNS-Optimization ($p{=}0.57$). Against Raw CoT, the backbone's own reasoning, CRAFT is significantly better in seven of the eight cells; the eighth is FLD with Gemini ($p{=}0.29$).

\begin{table*}[t]
\centering
\setlength{\tabcolsep}{3pt}
\renewcommand{\arraystretch}{1.1}
{\footnotesize
\begin{tabular}{l c c c c}
\toprule
\textbf{Setting} & \textbf{FLD} & \textbf{ProofWriter} & \textbf{Omni-MATH} & \textbf{OlympiadBench} \\
\midrule
\rowcolor{gray!20}\multicolumn{5}{c}{\textbf{GPT-5.4-nano}} \\
\midrule
    Direct & 66.1 [61.9, 70.2] & 72.0 [67.9, 75.8] & 6.9 [5.0, 9.5] & 4.0 [2.6, 6.1] \\
    Raw CoT & 77.0 [73.1, 80.5] & 84.6 [81.2, 87.5] & 33.7 [29.7, 38.0] & 44.1 [39.8, 48.5] \\
    Best-of-N & 78.0 [74.2, 81.4] & 87.2 [84.0, 89.8] & 40.8 [36.6, 45.2] & 51.1 [46.7, 55.5] \\
    Self-Consistency & 78.6 [74.8, 82.0] & 87.0 [83.8, 89.7] & 48.8 [44.4, 53.2] & 57.6 [53.2, 61.8] \\
    Univ.\ Self-Consistency & 69.2 [65.0, 73.1] & 70.8 [66.7, 74.6] & 19.5 [16.3, 23.2] & 20.1 [16.8, 23.9] \\
    Self-Refine & 62.5 [58.2, 66.6] & 94.0 [91.6, 95.8] & 46.8 [42.4, 51.2] & 55.1 [50.7, 59.4] \\
    Self-Eval Beam Search & 60.1 [55.7, 64.3] & 67.6 [63.4, 71.6] & 18.5 [15.3, 22.2] & 22.5 [19.1, 26.4] \\
    Faithful CoT & 78.6 [74.8, 82.0] & 85.8 [82.5, 88.6] & 39.2 [35.0, 43.6] & 46.9 [42.5, 51.3] \\
    Tree-of-Thought & 72.0 [67.9, 75.8] & 79.4 [75.6, 82.7] & 37.0 [32.8, 41.3] & 47.7 [43.3, 52.1] \\
    RAP & 72.8 [68.7, 76.5] & 89.6 [86.6, 92.0] & 36.8 [32.6, 41.1] & 49.1 [44.7, 53.5] \\
    PNS-Optimization & 75.4 [71.4, 79.0] & 76.4 [72.5, 79.9] & 36.6 [32.5, 40.9] & 47.7 [43.3, 52.1] \\
    Self-Aggregation & 76.8 [72.9, 80.3] & 86.6 [83.3, 89.3] & 32.3 [28.3, 36.6] & 43.5 [39.2, 47.9] \\
\cdashline{1-5}
    \textbf{CRAFT (Ours)} & \textbf{83.0} [79.5, 86.1] & \textbf{96.2} [94.1, 97.6] & \textbf{55.5} [51.1, 59.8] & \textbf{63.2} [58.9, 67.3] \\
\midrule
\rowcolor{gray!20}\multicolumn{5}{c}{\textbf{Gemini-3.1-Flash-Lite}} \\
\midrule
    Direct & 74.8 [70.8, 78.4] & 62.4 [58.1, 66.5] & 17.5 [14.4, 21.1] & 18.9 [15.7, 22.6] \\
    Raw CoT & 87.5 [84.3, 90.1] & 61.0 [56.7, 65.2] & 45.1 [40.8, 49.5] & 61.6 [57.2, 65.7] \\
    Best-of-N & 89.1 [86.1, 91.6] & 60.8 [56.5, 65.0] & 46.1 [41.8, 50.6] & 61.2 [56.8, 65.4] \\
    Self-Consistency & 89.9 [87.0, 92.3] & 61.0 [56.7, 65.2] & 55.1 [50.7, 59.4] & 63.8 [59.5, 67.9] \\
    Univ.\ Self-Consistency & 80.7 [76.9, 83.9] & 52.8 [48.4, 57.1] & 24.0 [20.4, 27.9] & 29.4 [25.5, 33.5] \\
    Self-Refine & 89.9 [87.0, 92.3] & 68.6 [64.4, 72.5] & 54.7 [50.3, 59.0] & 65.8 [61.5, 69.8] \\
    Self-Eval Beam Search & 80.0 [76.3, 83.3] & 65.8 [61.5, 69.8] & 37.0 [32.8, 41.3] & 48.7 [44.3, 53.1] \\
    Faithful CoT & 86.7 [83.4, 89.4] & 58.2 [53.8, 62.4] & 46.8 [42.4, 51.2] & 59.0 [54.6, 63.2] \\
    Tree-of-Thought & 86.7 [83.4, 89.4] & 60.6 [56.3, 64.8] & 45.3 [41.0, 49.7] & 59.4 [55.0, 63.6] \\
    RAP & 86.5 [83.2, 89.2] & 63.6 [59.3, 67.7] & 49.6 [45.2, 54.0] & 61.2 [56.8, 65.4] \\
    PNS-Optimization & \textbf{90.3} [87.4, 92.6] & 59.4 [55.0, 63.6] & 49.4 [45.0, 53.8] & 63.2 [58.9, 67.3] \\
    Self-Aggregation & 86.3 [83.0, 89.0] & 64.0 [59.7, 68.1] & 49.0 [44.6, 53.4] & 61.8 [57.4, 65.9] \\
\cdashline{1-5}
    \textbf{CRAFT (Ours)} & 89.3 [86.3, 91.7] & \textbf{86.6} [83.3, 89.3] & \textbf{59.3} [55.0, 63.6] & \textbf{73.4} [69.4, 77.1] \\
\bottomrule
\end{tabular}}
\vspace{-6pt}
\caption{Accuracy (\%) from Table~\ref{tab:icl_main} with its 95\% Wilson score interval, computed from each cell's own $n$. The interval is not centered on the accuracy, so both bounds are given.}
\vspace{-6pt}
\label{tab:ci}
\end{table*}

\section{TF-IRF}
\label{app:tfirf}
For each term $w$ not on the \textsc{CommonLogicalWords} list~\citep{bird-etal-2009-nltk} we compute three quantities over the $K$ traces of a problem:
\begin{align}
\overline{\mathrm{TF}}(w) &= \frac{1}{K}\!\sum_{k=1}^{K}\frac{\#(w, t_k)}{\sum_{w'}\!\#(w', t_k)}, \label{eq:tf}\\[4pt]
\overline{\mathrm{RF}}(w) &= \frac{|\{s \in S : w \!\in\! s\}|}{|S|}, \label{eq:rf}\\[4pt]
\mathrm{TF\text{-}IRF}(w) &= \overline{\mathrm{TF}}(w)\!\cdot\!\log\frac{1}{\overline{\mathrm{RF}}(w)}, \label{eq:tfirf}
\end{align}
where $t_k$ is trace $k$, $\#(w, t_k)$ the count of $w$ in it, and $S$ the set of steps in the $K$ traces of the problem. The consensus term set combines the TF-IRF threshold $\alpha$ with the step frequency threshold $\beta$: $T_{\text{Con}}=\{w\mid\mathrm{TF\text{-}IRF}(w)>\alpha \;\wedge\; \overline{\mathrm{RF}}(w)\ge\beta\}$.

\section{Z-Score, Jaccard Similarity and Group Comparison}
\label{app:related_concepts}

\paragraph{Z-score.}
Each step $s$ of a trace gets a score $x_s$, its Jaccard overlap with the consensus terms. Within the $K$ traces of one problem we standardize the scores, $Z_s = (x_s - \mu)/\sigma$, with $\mu$ and $\sigma$ the mean and standard deviation over all steps of the group, and remove a step with $Z_s < \gamma$. The group is the unit because a problem's own traces set the scale: a mathematics problem and a logic problem have different vocabularies and different overlap levels, and a fixed absolute threshold would treat them alike.

\paragraph{Jaccard similarity.}
Two steps, or a step and the consensus term set, are compared as sets of normalized tokens, $\mathrm{Jaccard}(A,B) = |A \cap B| / |A \cup B|$, between 0 and 1. It sees shared words and not paraphrase, which is why the edge weight pairs it with the model's own confidence; the ablation's embedding variant replaces it with the cosine similarity of sentence embeddings.

\paragraph{Group-relative comparison.}
Following GRPO~\cite{Guo_2025}, a trace or a step is judged relative to the other members of its group, the $K$ traces sampled for the same problem, and not on an absolute scale. A step is anomalous when it is far from what the group does, which is what the z-score above measures; an edge is reliable when its weight summed over the group reaches a fixed fraction of $K$, which is what $\theta$ measures.

\section{Related Work Details}
\label{Detailed Related Work}

\paragraph{Voting and selection over several traces.}
Self-Consistency~\cite{wang2023selfconsistencyimproveschainthought} and Universal Self-Consistency~\cite{chen2023universalselfconsistencylargelanguage} draw $K$ traces and vote on the final answer, treating each trace as a black box. Best-of-N~\cite{stiennon2022learningsummarizehumanfeedback} scores whole traces and returns the top one, and minimum Bayes risk selection~\cite{eikema2020mapdecodingneedinadequacy} returns the candidate closest to the others. All of them decide at the level of the final answer or of a score for the whole trace; none looks inside the steps. CRAFT's consensus RKG can be read as MBR applied to structure: instead of picking the candidate closest to the rest, it builds the structure the candidates share and writes a new trace along it.

\paragraph{Iterative self-refinement.}
Self-Refine~\cite{madaan2023selfrefineiterativerefinementselffeedback} has the model critique and rewrite one trace for several rounds; Self-Evaluation Guided Beam Search~\cite{xie2023selfevaluationguidedbeamsearch} grows a trace step by step under the model's own step scores. Both take the model's judgement of itself as the quality signal, so a confidently wrong model refines toward its error. CRAFT's signal is agreement among independent rollouts, which does not depend on the model judging its own output.

\paragraph{Recombining steps across chains.}
Recursive Self-Aggregation~\cite{venkatraman2026recursiveselfaggregationunlocksdeep} draws subsets of a population of chains, recombines their steps and repeats over rounds, so it does use the content of intermediate steps. It has no consensus graph and no filtering stage before recombination, so a step that only one chain contains is recombined like any other. CRAFT filters first, by edge weight and node support in the RKG, and synthesizes afterwards.

\paragraph{Graphs over a single trace.}
Graph-of-Thought~\cite{Besta_2024} and GraphCoT~\cite{jin2024graphchainofthought} arrange the reasoning of one generation episode as a graph, which allows backtracking and local refinement. The graph is built from one trace, so there is no second trace to compare it with and no way to see that a step is unusual. CRAFT's graph is aggregated across $K$ traces, and the counts that aggregation produces are the statistics its filters use.

\paragraph{Graphs over several chains with selection.}
MGRS~\cite{yang2025mgrs} builds a directed acyclic graph over several candidate chains, estimates the success rate of intermediate nodes by cross-chain verification, and returns the original chain with the best cumulative score. The verification goes further than voting, but the output is still one of the input chains. CRAFT returns a trace that did not exist among the inputs, generated node by node in the topological order of the consensus graph, and uses the same graph for filtering and for generation.

\paragraph{Step-level intervention by training.}
ConCISE~\cite{qiao-etal-2025-concise} fine-tunes a reasoning model to suppress two patterns of redundant reflection and so shortens its traces. DICE~\cite{li2025dicestructuredreasoningllms} trains a small model to post-process a large model's chain into a required structure. PNS-Optimization~\cite{yu2025causalsufficiencynecessityimproves} scores each step's necessity and sufficiency for the answer and filters training data by it; as a baseline we run only its training-free step scoring (Appendix~\ref{app:baselines}). Each of the three original methods needs a training stage and addresses one kind of flaw; CRAFT is training-free and its consensus filter catches both \emph{Step Internal Flaws} and \emph{Step-wise Flaws}.

\paragraph{Process reward models and auditors.}
Process reward models~\cite{lightman2023letsverifystepstep} are step verifiers trained on human labels and applied to one trace at a time; AgentAuditor~\cite{yang2026agentauditor} audits agent reasoning with an LLM judge. Both add a verifier model. CRAFT uses no verifier; a step is suspect when it departs from what the other rollouts did. The two approaches can be combined, for example by using a PRM to score CRAFT's synthesized traces.

\paragraph{Debate.}
Multi-agent debate~\cite{du2024improvingfactualityreasoninglanguage} has several model instances argue toward a conclusion. In debate the instances exchange claims, while CRAFT compares step graphs and the traces never see each other. Debate could follow synthesis as a verification stage~\cite{wang2026specalignefficientspecificationgroundedalignment}.

\section{Case Study}
\label{app:case_study}

Two traces from our runs show the two flaw types of Figure~\ref{fig:problem background}. The first is a FOLIO problem from our preliminary runs, the second an OlympiadBench problem. Both reach the right answer.

\smallskip\noindent
\begin{tabular}{@{}p{1.0\columnwidth}@{}}
\toprule
\textbf{Example 1} \\
\midrule
Step~1: From Fact~1, TS~Leda was a good passenger and cargo vessel. \hfill \textcolor{green!60!black}{\ding{51}} \\
Step~2: From Fact~4, TS~Leda must be quiet and good at sea. \hfill \textcolor{green!60!black}{\ding{51}} \\
Step~3: From Fact~5, \emph{some} quiet/seaworthy ships have powerful steam turbine engines. Using Step~2, this implies TS~Leda has powerful steam turbine engines. \hfill \textcolor{red}{\ding{55}} \emph{quantifier fallacy ($\exists \!\to\! \forall$)} \\
Step~4: From Fact~3, stabilizers are found only on ships with powerful steam turbine engines. Since TS~Leda has such engines (Step~3), it can have stabilizers. \hfill \textcolor{red}{\ding{55}} \emph{propagates Step~3 error} \\
Step~5: From Fact~2, TS~Leda was a Norwegian vessel with stabilizers. \hfill \textcolor{green!60!black}{\ding{51}} \\
Step~6: Hypothesis says not Norwegian, but Step~5 (Fact~2) directly contradicts this. \hfill \textcolor{green!60!black}{\ding{51}} \\
\textbf{Label: \textsc{disproved}} \hfill \textcolor{green!60!black}{\ding{51}}\\
\bottomrule
\end{tabular}

\subsection*{Example 1: a Step Internal Flaw}

\emph{Problem:} Fact~1: TS~Leda was a good passenger and cargo vessel. Fact~2: TS~Leda was a Norwegian vessel that was built with stabilizers. Fact~3: Stabilizers are found only on ships with powerful steam turbine engines. Fact~4: To be a good passenger/cargo vessel, ships must be quiet and good at sea. Fact~5: \emph{Some} ships that are quiet and good at sea have powerful steam turbine engines. Fact~6: Vessels are ships. \textbf{Hypothesis:} TS~Leda was \emph{not} a Norwegian vessel. \textbf{Ground truth: \textsc{disproved}.}

The label is right and Step~3 is wrong. Fact~5 says that \emph{some} quiet, seaworthy ships have such engines, and the step concludes that TS~Leda in particular does; that is an existential read as a universal, and Step~4 builds on it. The trace still lands on \textsc{disproved} because Step~5 goes back to Fact~2, which settles the hypothesis by itself, so Steps~3 and~4 do not affect the label here. A student fine-tuned on this trace would learn the inference in Step~3 as a valid move. In CRAFT the two steps are the kind that consensus removes: other rollouts of the same problem do not derive the engines, so the step's terms are absent from the consensus set and its edge into Step~4 has low support.

\smallskip\noindent
\begin{tabular}{@{}p{1.0\columnwidth}@{}}
\toprule
\textbf{Example 2} \\
\midrule
Step~1: Write $N=10a+b$; define $R(M)$ as the digit-reversal of $M$. Need $R(N)$, $R(4N)$, $R(5N)$ all prime. \hfill \textcolor{green!60!black}{\ding{51}} \\
Step~2: Since $N\le99$, note $5N\le495$ (three digits). Primes cannot end in 4, so $R(5N)$ must not end in 4. \hfill \textcolor{green!60!black}{\ding{51}} \\
Step~3: Digit parity: for $R(N)$ to be prime and $>5$, its units digit must be odd, so the tens digit $a$ of $N$ is odd: $a\in\{1,3,5,7,9\}$. \hfill \textcolor{green!60!black}{\ding{51}} \\
Step~4: Similarly restrict $4N$ and $5N$ by parity of their reversal's units digit. \hfill \textcolor{green!60!black}{\ding{51}} \\
Step~5: Try $N=99$: $R(99)=99=9\times11$, not prime. \hfill \textcolor{orange}{\ding{55}} \emph{redundant} \\
Step~6: Try $N=97$: $5\times97=485$, $R(485)=584$, even --- not prime. \hfill \textcolor{orange}{\ding{55}} \emph{redundant} \\
Steps~7--23: Try $N=96,95,94,\ldots,80$ one by one (each fails $\ge1$ condition). \hfill \textcolor{orange}{\ding{55}} \emph{redundant $\times$17} \\
Step~24: Try $N=79$: $R(79)=97$\,(prime), $4\times79=316$, $R(316)=613$\,(prime), $5\times79=395$, $R(395)=593$\,(prime). \hfill \textcolor{green!60!black}{\ding{51}} \\
Step~25: Conclude $N=79$ is the largest. \hfill \textcolor{green!60!black}{\ding{51}} \\
\textbf{Answer: $N=79$} \hfill \textcolor{green!60!black}{\ding{51}} \\
\bottomrule
\end{tabular}

\subsection*{Example 2: a Step-wise Flaw}

\emph{Problem:} A \emph{reverse prime} is a positive integer $N$ whose digit reversal is prime (16 is one, since 61 is prime). Find the largest two-digit $N$ such that $N$, $4N$ and $5N$ are all reverse primes. \textbf{Ground truth: $N = 79$.}

Every step is arithmetically correct and the answer is right; the flaw is in how the steps are organized. Steps~1--4 derive constraints that cut the search down to a handful of candidates, and then Steps~5--23 ignore them and try every integer from 99 down to 80. Nineteen steps do work the first four had already made unnecessary; the short route is the bound $5N<400 \Rightarrow N\le79$ and three primality checks. Under CRAFT the enumeration steps are the ones that go: each carries tokens no other rollout uses (``try $N=96$'', ``try $N=95$''), so their TF-IRF weight is low and their edges appear in one trace out of $K$.

\section{LLM Usage Statement}
GPT-5.2 helped with grammar. Claude-4.6-Sonnet helped write and organize the code. For the unsupervised comparison in the method we asked Claude-4.6-Sonnet to choose among ideas we had drafted, and then refined the chosen one ourselves.

The API we used is provided by a partner company that, for data-security reasons, has the model providers deploy their models on its own GPU clusters. Performance through that endpoint can differ from the public APIs, and the numbers in this paper are from that endpoint.

\clearpage
\onecolumn

\section{TF-IRF Word Lists}

\begin{tcolorbox}[title=TF-IRF Word Lists]
\textbf{LOGICAL\_KEYWORDS}:
\begin{itemize}
    \item \textbf{Step-related}: step, steps, step1, step2, step3, step4, step5, first, second, third, next, then, finally.
    \item \textbf{Conditional}: if, else, elif, when, whenever, unless, provided.
    \item \textbf{Causality}: because, since, as, due, therefore, thus, hence, so, consequently, accordingly.
    \item \textbf{Connectors}: and, or, but, however, moreover, furthermore, additionally, also, besides.
    \item \textbf{Inference}: implies, imply, conclude, conclusion, infer, inference, deduce, deduction.
    \item \textbf{Proof-related}: prove, proof, proven, disprove, disproven, contradiction, contradictory, assume, assumption.
    \item \textbf{Fact-referencing}: fact, facts, given, premise, premises, hypothesis, statement, claim.
    \item \textbf{Logical operations}: all, any, some, none, every, each, not, no, never, always.
    \item \textbf{Comparison}: equal, equals, equivalent, same, different, greater, less, than, from.
\end{itemize}

\textbf{\textsc{CommonLogicalWords}}:

{\footnotesize because, also, so, and, or, but, if, then, therefore, thus, since, as, when, not, no, all, any, some, each, every, fact, facts, step, steps, prove, proof, conclude, conclusion, approve, disprove, given, statement, claim, implies, however.}

\textbf{STOPWORDS}:

{\footnotesize the, a, an, is, are, was, were, be, been, being, have, has, had, do, does, did, will, would, could, should, may, might, must, can, this, that, these, those, it, its, they, them, their, there, here, where, what, which, who, whom, whose, how, why, when, to, of, in, on, at, by, for, with, from, as, into, onto, up, down, out, off, over, under, above, below, between, among, through, during, before, after, while, about, against, within, without, throughout, across, around, near, far, inside, outside, beside, besides, except, including, excluding, concerning, regarding, according, i, you, he, she, we, us, our, your, my, me, him, her, his, hers, mine, yours, ours, theirs.}
\end{tcolorbox}

\section{Prompt Templates}
\label{Prompt Templates}

\subsection{Module I: Multi-Trace Generation}

\begin{tcolorbox}[title={System prompt, logical}, colback=blue!5!white, colframe=blue!50!black]
You are a meticulous logician. Produce exhaustive, atomic reasoning. Each step must cite the facts or earlier steps it depends on. Avoid circular references and avoid skipping steps.
\end{tcolorbox}

\begin{tcolorbox}[title={User prompt, logical (FLD and ProofWriter)}]
Generate extremely detailed reasoning for the following problem.

\textbf{Problem Statement:}
\{problem\}

\textbf{Instructions:}
\begin{itemize}
    \item Provide 6--10 numbered steps in the format ``Step 1:'', ``Step 2:'', etc.
    \item Each step must cite the exact facts or previous steps it uses.
    \item Use natural language sentences without JSON or markdown code fences.
    \item After the steps, include a short summary paragraph.
    \item Do not stop early; if the answer is cut off, immediately continue until the summary and final conclusion are delivered.
    \item Determine whether the hypothesis is \texttt{\_\_PROVED\_\_} or \texttt{\_\_DISPROVED\_\_} based on your reasoning.
    \item IMPORTANT: Your response is incomplete without a final conclusion line. You MUST end with EXACTLY this format: \texttt{Final Conclusion: \_\_PROVED\_\_} (or \texttt{\_\_DISPROVED\_\_} depending on your reasoning).
    \item A response missing the Final Conclusion line will be rejected and retried.
\end{itemize}
\end{tcolorbox}

\begin{tcolorbox}[title={System prompt, mathematical}, colback=blue!5!white, colframe=blue!50!black]
You are a meticulous mathematician. Produce exhaustive, step-by-step mathematical reasoning. Each step must show the complete transformation with full equations written out. Cite which previous result or given value you are using in each step. Never skip algebraic steps or arithmetic transitions.
\end{tcolorbox}

\begin{tcolorbox}[title={User prompt, mathematical (Omni-MATH and OlympiadBench)}]
Solve the following math problem with detailed step-by-step reasoning.

\textbf{Problem:}
\{problem\}

\textbf{Instructions:}
\begin{itemize}
    \item Provide 4--10 numbered steps in the format ``Step 1:'', ``Step 2:'', etc.
    \item In each step, write out the full equation or expression on BOTH sides (e.g., `$2x + 3 = 7 \rightarrow 2x = 4$').
    \item Each step must state which fact, given value, or previous result it uses.
    \item Do NOT skip arithmetic or algebraic manipulations.
    \item Use natural language sentences; do NOT use JSON or markdown code fences.
    \item Do not stop early; continue until you reach a numeric or symbolic final answer.
    \item The final step MUST state the answer clearly using \verb|\boxed{<answer>}| notation.
    \item End with a single line exactly formatted as: \texttt{Final Answer: \textbackslash boxed\{<answer>\}}
\end{itemize}
\end{tcolorbox}

\subsection{Module II: Graph Construction}

\begin{tcolorbox}[title=Dependency extraction prompt, colback=green!5!white, colframe=green!50!black, breakable]
You are analyzing a step-by-step reasoning trace. Your task is to identify the DIRECT dependencies between steps.

For each reasoning step, identify which facts or EARLIER steps it DIRECTLY depends on to reach its conclusion.
\begin{itemize}
    \item Only list DIRECT dependencies (not transitive ones).
    \item A step depends on another step if it uses that step's conclusion as a premise.
    \item Do NOT list a step as depending on itself.
    \item If a step uses no earlier steps or facts, set ``uses'' to [].
    \item For every dependency, give your confidence in [0, 1] that the step really uses it: 1.0 when the step states or plainly applies that premise, lower when the link is only implied, and do not list a dependency you would put below 0.3.
\end{itemize}

Output ONLY valid JSON with this exact structure:

{\scriptsize\texttt{\{"dependencies": [\{"step\_id": "Step1", "uses": []\}, \{"step\_id": "Step2", "uses": [\{"id": "Fact1", "confidence": 1.0\}, \{"id": "Step1", "confidence": 0.8\}]\}, \{"step\_id": "Step3", "uses": [\{"id": "Step2", "confidence": 0.9\}]\}]\}}}

Available facts (given premises): \{facts\_block\}

Reasoning steps: \{steps\_block\}

Respond with ONLY the JSON, no other text.
\end{tcolorbox}

\subsection{Module III: Topology-guided Trace Synthesis}

\begin{tcolorbox}[title={Synthesis prompt for one node, logical (FLD and ProofWriter)}, colback=green!5!white, colframe=green!50!black, breakable]
You are an expert logician building a rigorous reasoning chain.

\textbf{Problem}: \{problem\}

\textbf{Your current task}: Generate \{node\_id\} (Step \{i\} of \{n\})

\textbf{Previously Generated Steps}: \{steps written so far\}

(Direct prerequisites for this step: \{predecessor node ids\})

\hfill {\footnotesize For the first step, which has no steps before it, this reads ``\textbf{Direct prerequisites}: (none --- use only the problem statement above)''.}

\textbf{Reference step} (distilled from high-frequency traces --- \{p\}\% of the traces agree on it --- use as a quality anchor): ``\{consensus node text\}'' (Preserve the logical structure and content; you may rephrase for clarity.)

\hfill {\footnotesize If at most 40\% of the traces contain the node, the reference is instead introduced as ``only \{p\}\% of the traces derived this --- treat it as a suggestion, not a settled result'' and the model is told to derive the step itself and keep it only if it follows.}

\textbf{Key terms for this step} (frequency-weighted from \{k\} traces): \{terms\}

\textbf{Requirements}:
\begin{itemize}
    \item Build this step on the prerequisites named above.
    \item ONE inference only: apply exactly one rule to exactly one set of premises and state what follows. Do not chain two rules in this step, and do not restate what earlier steps established.
    \item \emph{(Intermediate steps.)} Say the rule and its application in TWO sentences: first the rule or fact in the problem's own words, then what follows when it is applied to the premises named.
    \item \emph{(Intermediate steps.)} Write that first sentence as your own sentence carrying the rule --- ``According to Fact 12, \ldots'' --- not as a quotation dropped in after a colon.
    \item \emph{(Final step.)} End this step with the hypothesis in the problem's own wording and the verdict on it. Say whatever this step still needs to establish first.
    \item If the rule carries nested negation (``\ldots\ does not hold'', ``it is not \ldots''), first restate it in the affirmative and say what it reduces to, then apply the restated form.
    \item \emph{(Intermediate steps.)} Do NOT conclude the entire problem here --- more steps follow.
    \item \emph{(Final step.)} This is the FINAL step --- you MUST end with the EXACT string \texttt{\_\_PROVED\_\_} or \texttt{\_\_DISPROVED\_\_} as the last part of your response. Example final sentence: ``Therefore, the hypothesis is \texttt{\_\_PROVED\_\_}.'' Do NOT omit this marker --- it is mandatory.
    \item \emph{(Final step.)} \texttt{\_\_PROVED\_\_} means the hypothesis AS WRITTEN follows from the facts; \texttt{\_\_DISPROVED\_\_} means its negation follows. Before choosing, restate the hypothesis with whatever negation it carries and check it against what you derived: deriving ``X does not hold'' PROVES a hypothesis that says X does not hold, and DISPROVES one that says X holds.
    \item \emph{(Final step.)} Write that restatement out in the problem's own wording, in this final step, before the marker.
    \item \emph{(Final step.)} Prior: the independent reasoning traces mostly reach \{majority label\} (\{s\}\% of the weighted vote). This is a prior, not the answer --- the traces come from one model and can agree on a mistake. Decide from the chain derived above: if it supports \{majority label\}, say so; if it supports the other label, state the other label instead.
\end{itemize}

\hfill {\footnotesize In the configurations that follow the vote, the prior line instead reads ``The majority of independent reasoning traces reach \{majority label\}. Generate reasoning that rigorously leads to this conclusion.''}

Generate ONLY \{node\_id\}: [your reasoning here]
\end{tcolorbox}

\begin{tcolorbox}[title={Synthesis prompt for one node, mathematical (Omni-MATH and OlympiadBench)}, colback=green!5!white, colframe=green!50!black, breakable]
You are an expert mathematician. Your task is to solve the math problem with detailed, step-by-step calculations based on the important terms extracted from multiple reasoning traces.

\textbf{Problem Context}: \{problem\}

\textbf{Prior} (the majority of the independent traces reached this; they come from one model and can agree on a mistake --- treat it as a prior, not as the answer, and follow your own derivation if it disagrees): \{majority answer\}

\textbf{Previously Generated Steps}: \{steps written so far\}

\textbf{Current Step (\{i\}) - Important Terms} (extracted from \{k\} traces): Key terms: \{terms\}; Term frequencies: \{term counts\}

\textbf{Critical Quality Guidelines} (MUST FOLLOW):
\begin{itemize}
    \item \textbf{Show the Work That Matters}: write the equation being solved and its result. Arithmetic that a reader can do in their head need not be spelled out.
    \item \textbf{Say It In Two Sentences}: write the prose of this step as two sentences, three at the most. Where it would take several short ones, join them; never leave out a calculation to meet the count --- if the step needs an intermediate value, write it.
    \item \textbf{Stand On The Problem's Words}: name the given condition this step uses as the problem states it, before applying it.
    \item \textbf{Cite Premises}: Each step must state which given value, equation, or previous result it uses.
    \item \textbf{No Arithmetic Errors}: Double-check every calculation. If uncertain, re-derive from the previous step.
    \item \textbf{No Hallucination}: Use ONLY values and equations present in the Problem Context or derived in previous steps.
    \item \textbf{No Redundancy}: Each step must advance the solution. Do not restate already-established equations.
    \item \textbf{Reach a Numeric Answer} (CRITICAL): You MUST continue until you obtain a concrete numeric or symbolic answer. The final step MUST include the answer in \verb|\boxed{<answer>}| notation. Do NOT stop before boxing the answer.
\end{itemize}

\textbf{Current Task}: Generate Step \{i\} of \{n\} total steps \hfill {\footnotesize (final step adds ``(THIS IS THE FINAL STEP - MUST REACH A CLEAR CONCLUSION)'')}

\textbf{CRITICAL REQUIREMENTS}:
\begin{itemize}
    \item \textbf{Logical Consistency Check}: CAREFULLY review ALL previously generated steps above. Identify what conclusions/facts have been established in previous steps. Use ONLY the established conclusions from previous steps as premises. Ensure your reasoning logically follows from these established facts. If a previous step concluded ``X'', you can use ``X'' as a premise in this step. DO NOT contradict or ignore conclusions from previous steps.
    \item \textbf{Step Generation}: Generate ONLY Step \{i\} based on the important terms provided. Build logically on the previously generated steps (use their conclusions as premises). Use the key terms naturally in your reasoning. Follow all quality guidelines above. Output format: ``Step \{i\}: [your reasoning here]''. Do NOT generate any other steps.
    \item \emph{(Intermediate steps.)} \textbf{Continuation Requirements}: This is Step \{i\} of \{n\} --- you are NOT done yet. Continue building the reasoning chain. Do NOT conclude or stop here. Prepare for the next step.
    \item \emph{(Final step.)} \textbf{Final Step Requirements} (CRITICAL --- This is Step \{i\}, the LAST step): You MUST arrive at a clear, numeric or symbolic answer. Express the final answer using \verb|\boxed{<answer>}| notation. Do NOT stop before boxing the answer. Example: ``Therefore, $x = 3$, so the answer is \verb|\boxed{3}|''. This is REQUIRED --- do not skip the boxed answer.
\end{itemize}

\textbf{Before Generating, Verify}:
\begin{itemize}
    \item Does this step make exactly one inference?
    \item Does it state something the steps before it have not?
    \item Is its prose two sentences, three at the most?
    \item Is every intermediate value this step relies on written down?
    \item If this is the final step, will you reach an explicit conclusion?
\end{itemize}

Please generate Step \{i\} now:

\textbf{Consensus node for this step} (what the surviving traces agreed this step establishes): ``\{consensus node text\}'' Reach that result in this step. Treat the wording as the target, not as a source to copy: re-derive every number yourself, and where your own arithmetic disagrees with it, follow your arithmetic and say so.

\emph{(Final step.)} Note: majority of traces suggest the answer may be \{majority answer\} --- verify against your computation. This is the FINAL step --- compute the exact final answer. Must end with \verb|\boxed{<answer>}|.
\end{tcolorbox}

\subsection{Pilot Study: Trace Generation for ROSCOE}

\begin{tcolorbox}[title=Reading comprehension (CosmosQA and DROP), colback=orange!5!white, colframe=orange!60!black]
\textbf{System.} You are an expert at reading comprehension and reasoning. Given a passage and a question, reason step-by-step to find the answer. Each step must be a single sentence. End with a final sentence stating the answer.

\textbf{User.} Passage and question: \{passage and question\} \\
\textit{Correct answer: \{answer\}} \hfill {\footnotesize (w/ Answer only)} \\
Write a numbered step-by-step reasoning chain to answer the question. End with a final sentence stating the answer.
\end{tcolorbox}

\begin{tcolorbox}[title=Natural language inference (eSNLI), colback=orange!5!white, colframe=orange!60!black]
\textbf{System.} You are an expert at natural language inference. Given two sentences (premise and hypothesis), reason step-by-step to determine whether the hypothesis is entailed, contradicted, or neutral with respect to the premise. Each step must be a single sentence. End with a final sentence stating your conclusion (entailment / contradiction / neutral).

\textbf{User.} Premise: \{premise\} \\ Hypothesis: \{hypothesis\} \\
\textit{Correct answer: \{entailment / contradiction / neutral\}} \hfill {\footnotesize (w/ Answer only)} \\
Write a numbered step-by-step reasoning chain to determine the relationship (entailment / contradiction / neutral) between premise and hypothesis. End with a final sentence stating your conclusion.
\end{tcolorbox}

\begin{tcolorbox}[title=Math word problems (GSM8K), colback=orange!5!white, colframe=orange!60!black]
\textbf{System.} You are an expert at solving math word problems. Given a math problem, solve it step by step. Each step must be a single sentence showing one calculation or logical deduction. End with a final sentence stating the numeric answer.

\textbf{User.} Problem: \{problem\} \\
\textit{Correct answer: \{answer\}} \hfill {\footnotesize (w/ Answer only)} \\
Write a numbered step-by-step solution. End with a final sentence stating the numeric answer.
\end{tcolorbox}

\subsection{Pilot Study: PRMBench Step Verifier}
\label{app:prmbench_prompts}

\begin{tcolorbox}[title=System prompt (w/ Answer; the w/o Answer form omits ``its correct final answer{,}''), colback=orange!5!white, colframe=orange!60!black, breakable]
You are a mathematical reasoning verifier.
You will be given a math problem, its correct final answer, and numbered solution steps.
For each step, output two scores:
\begin{itemize}
    \item \textbf{validity}: $+1.0$ if the step is logically correct, $-1.0$ if it contains an error, or a value in between for partial errors.
    \item \textbf{redundancy}: $+1.0$ if the step is redundant (adds no new information), $-1.0$ if it is necessary and informative.
\end{itemize}
Respond with ONLY a JSON object: \texttt{\{"validity": [s1, s2, \ldots], "redundancy": [s1, s2, \ldots]\}}

\medskip
\textbf{Example:}\\
Question: What is 15\% of 80?\\
Solutions:\\
Step 1: To find 15\% of 80, we convert the percentage to a decimal: $15\% = 0.15$.\\
Step 2: Now multiply: $0.15 \times 80 = 120$.\\
Step 3: Therefore the answer is 120.

Output: \texttt{\{"validity": [1.0, -1.0, -1.0], "redundancy": [-1.0, -1.0, -1.0]\}}

Explanation: Step 1 is valid. Step 2 has an arithmetic error ($0.15 \times 80 = 12$, not 120), so validity $= -1$. Step 3 propagates the wrong answer, so validity $= -1$.
\end{tcolorbox}

\begin{tcolorbox}[title=User prompt, colback=orange!5!white, colframe=orange!60!black, breakable]
Question: \{question\}

\textit{Correct final answer: \{correct\_answer\}} \hfill {\footnotesize (w/ Answer only)}

Solutions:\\
Step 1: \{step\_1\}\\
Step 2: \{step\_2\}\\
$\vdots$

Output a JSON object with validity and redundancy arrays, each of length \{n\_steps\}.
\end{tcolorbox}

\end{document}